# Towards geological inference with process-based and deep generative modeling, part 1: training on fluvial deposits

Guillaume Rongier[1] and Luk Peeters[2]

[1]*Department of Geoscience & Engineering, Delft University of Technology, Delft, The Netherlands*
[2]*VITO Digital Water & Soils, Mol, Belgium*

**Abstract** The distribution of resources in the subsurface is deeply linked to the variations of its physical properties. Generative modeling has long been used to predict those physical properties while quantifying the associated uncertainty. But current approaches struggle to properly reproduce geological structures, and fluvial deposits in particular, because of their continuity. This study explores whether a generative adversarial network (GAN) – a type of deep-learning algorithm for generative modeling – can be trained to reproduce fluvial deposits simulated by a process-based model – a more expensive model that mimics geological processes. An ablation study shows that developments from the deep-learning community to generate large 2D images are directly transferable to 3D images of fluvial deposits. Training remains stable, and the generated samples reproduce the non-stationarity and details of the deposits without mode collapse or pure memorization of the training data. Using a process-based model to generate those training data allows us to include valuable properties other than the usual physical properties. We show how the deposition time let us monitor and validate the performance of a GAN by checking that its samples honor the law of superposition. Our work joins a series of previous studies suggesting that GANs are more robust that given credit for, at least for training datasets targeting specific geological structures. Whether this robustness transfers to larger 3D images and multimodal datasets remains to be seen. Exploring how deep generative models can leverage geological principles like the law of superposition shows a lot of promise.

## 1 Introduction

Sedimentary deposits from rivers represent an invaluable capital for our society, either directly by providing building materials or indirectly by the resources they can host. These include key commodities to adapt to and mitigate climate change: groundwater, geothermal heat, metals, hydrogen, or $CO_2$ (e.g., Morris & Ramanaidou, 2007; Donselaar et al., 2017; Zhou et al., 2020). When factoring in population growth, these fluvial deposits will become more densely exploited in the future. It raises the question of their sustainable management, especially considering that some applications will be in competition for the same domain. This sustainable management is impossible without characterizing the spatial variability of those deposits and the variability of their physical properties, such as porosity and permeability, which control where we can find or store resources. Unfortunately, subsurface data are too scarce or poorly resolved to capture those properties at a sufficient level of detail for decision-making. This explains the longstanding use of generative modeling to predict the physical properties of the subsurface while quantifying uncertainties.

The generative models at the heart of geological modeling are data-driven and based on a weak geological prior (Deutsch & Journel, 1992), i.e., a mean and a covariance function that neglect the high-order moments essential to reproduce geological structures (Guardiano & Srivastava, 1993; Gómez-Hernández & Wen, 1998; Journel, 2005). A weak prior is not an issue in the high-data context often seen in machine learning, where similar approaches are used (Rasmussen & Williams, 2006), but it is one in the low-data context of a non-stationary subsurface. This led to the development of a multi-step modeling strategy in which facies, i.e., clusters of rocks with similar physical properties, and trends are identified and modeled first to improve the geological plausibility of the predicted physical properties (e.g., Deutsch & Wang, 1996; W. Xu, 1996; Strebelle, 2002). This manual feature engineering is time-consuming, and the resulting workflow cumbersome. As a result, subsurface models are often based on a single conceptual model – i.e., an internally consistent summary of our understanding of a geological system – missing a major contribution to uncertainty (Enemark et al., 2019).

The limitations of this strategy come partly from only considering the current state of the subsurface, and ignoring how it came to be. Fluvial deposits are not distributed randomly, but stem from the physical processes behind water runoff and sediment transport (Bridge, 2007). By simulating those processes, process-based models (e.g., Tetzlaff & Harbaugh, 1989; Granjeon & Joseph, 1999; Clevis et al., 2004) offer us a straightforward way to generate realizations consistent with geological principles, improving the geological plausibility of our predictions. They also allow us to go beyond facies, which are more an interpretation than a measurable quantity. Instead, we can simulate the distribution of different grain sizes of sediments. Such properties are closer to real measurements, capture heterogeneities in more detail, and have a direct, physical link to other properties of interest, such as porosity and permeability. But process-based models are hard to use in practice because of their computational cost and the difficulty in inferring their initial and boundary conditions, e.g, initial topography and rainfall.

Deep generative modeling (Tomczak, 2022) appears as an attractive solution owing to its exceptional progress in generating all kinds of images over the last decade (e.g., Goodfellow et al., 2014; Karras et al., 2018; Brock et al., 2019). In this context, it acts as an emulator: a model based on deep neural networks learns to generate fluvial deposits from realizations simulated by a process-based model. Once trained, the deep generative model can generate new realizations much faster than the process-based model. An essential benefit from deep learning resides in its aptitude for representation learning (Bengio et al., 2013), i.e., its aptitude to extract hierarchical information from data that makes a specific task easier. While geostatistical approaches require some manual feature engineering to capture the non-stationarity of the subsurface, deep learning can do the same automatically given enough training samples. While deep learning's need for large training datasets is often perceived as a weakness, it can actually become an asset: by using training datasets covering a large range of sedimentary settings, we can reduce conceptual biases.

In this article, we aim to uncover the applicability, benefits, and limitations of combining deep generative modeling with process-based modeling to generate fluvial deposits. We focus on a specific type of deep generative model, a generative adversarial network (GAN), and on three main questions:

1. Can a GAN generate samples representing plausible fluvial deposits based on continuous, non-stationary properties?

2. Can we consistently train GANs for 3D image generation based on existing architectures and limited tuning?

3. Can we use geological principles to increase the practical value of GANs and better assess their performance?

## 2 Materials and methods

This section provides a general overview of the approaches used in this work. We refer readers who are after more details to the articles cited in the subsections below and to the implementation that supports our study (Rongier & Peeters, 2025b).

### 2.1 Training and testing data

Our training and testing data come from FluvDepoSet, a dataset of 20 200 synthetic 3D realizations of fluvial deposits (Rongier & Peeters, 2021). Each realization was simulated using the Channel-Hillslope Integrated Landscape Development Model (CHILD) (Tucker, Lancaster, Gasparini, Bras, & Rybarczyk, 2001; Tucker, Lancaster, Gasparini, & Bras, 2001). CHILD is a landscape evolution model that uses empirical and simplified physical laws to reproduce efficiently the wide range of processes driving landscape evolution. All realizations started from a flat, titled topography (Rongier & Peeters, 2025a). Water entered the system through an inlet in the middle of the upper side of the domain and through rainfall all over the domain following a stochastic process. It left the system through an outlet in the middle of the lower side. Between the inlet and outlet, a river formed then migrated laterally following a topographic steering mechanism. The elevation of the inlet could vary up and down, depicting externally imposed aggradation and incision phases. Stratigraphy was built based on two grain sizes: a coarse fraction of sediments was deposited in point bars along the river, while a fine fraction was deposited in the overbank during flood events. This stratigraphy was initially stored into an irregular structured grid that followed the topography. At the end of the simulation, it was transferred into a regular structured grid so that all realizations share a common grid. This grid has $128 \times 200 \times 32$ cells, with each cells being $50 \times 50 \times 0.5$ m large, and contains two properties of interest: the fraction of coarse sediments and the deposition time of the sediments. CHILD's input parameters remained the same between all the realizations except for seven parameters randomly drawn from uniform distributions: the bank erodibility, the river inlet elevation through time, the overbank distance decay constant, the overbank deposition rate constant, the coarse grain diameter, the fine grain diameter, and the mean storm rainfall through time (see Rongier & Peeters, 2025a, for more details).

Extracting a sample suitable for training and testing from a realization follows four steps repeated each time a realizations is used (examples of such samples can be found in figures 2 and 3):

1. The realization is cropped along the vertical axis (between 4 and 14 m) to remove the initial substratum and most of the space above the final topography, so to focus on the sediments deposited during CHILD's run.

2. Some areas still contain some empty space above the final topography. In those areas, the fraction of coarse sediments is filled with zeros (fine sediments only) and the deposition time of the sediments is filled with the age of the sediments directly below plus 1 year.

3. GANs are usually trained with squared images of size in powers of 2, so the cropped realization is cropped further to extract a sample of $128 \times 128 \times 16$ cells. This second cropping can be based on specific coordinates – so deterministic, like the first one – or random. A random cropping returns a slightly different sample each time the realization is used, which may help to stabilize training, akin to data augmentation. But in some cases, for instance to test for memorization (see section 3.2), a deterministic cropping is more valuable.

4. The fraction of coarse sediments and the deposition time of the sediments are scaled between -1 and 1. The fraction of coarse sediments is scaled based on its absolute boundaries (0 and 1), while the deposition time is scaled independently for each realization.

The realizations 1 to 20 000 are used in the training process, with 19 000 realizations used for the training itself and 1 000 used for validation. The division between training and validation set is random, except when exploring the impact of the number of training samples, where realizations 19 001 to 20 000 are used for validation (see section 3.3). The realizations 20 001 to 20 200 are kept for testing.

## 2.2 Generative adversarial networks

A generative adversarial network is a type of deep generative model based on two deep neural networks: a discriminator ($D$) and a generator ($G$) (Goodfellow et al., 2014). The discriminator aims at telling whether a sample is real – i.e., it comes from the training data – or fake. The generator aims at generating fake samples that fool the discriminator. The generator takes as input some random noise ($z$) that comes from a latent variable ($Z \sim \mathcal{N}(0,1)$) and returns a fake sample ($G(z)$). The discriminator takes as input a real or fake sample ($x$ or $G(z)$ respectively) and returns the probability of the sample being real ($D(x)$ or $D(G(z))$). Training the discriminator means maximizing the probability that a sample from the training data is labeled as real and a sample from the generator as fake. Training the generator means minimizing the probability that a sample from the generator is labeled as fake. Training both networks together is a zero-sum game, so the goal is to find an equilibrium, called Nash equilibrium, through a two-player minimax game formalized as:

$$\min_G \max_D \mathbb{E}_{x \sim p_{\text{data}}(x)} \left[ \log D(x) \right] + \mathbb{E}_{z \sim \mathcal{N}(0,1)} \left[ \log \left( 1 - D(G(z)) \right) \right] \quad (1)$$

Finding this equilibrium is a notoriously difficult task. A common issue arises when one network overcomes the other, in which case both networks stop learning. But GANs also tend to suffer from mode collapse, in which case the generator only captures a subset of the training data.

Finding strategies to consistently obtain a stable training, no matter the training dataset, has been the main focus of research around GANs. Those strategies usually evolve around specific architectures for the generator and discriminator and regularization schemes for the minimax optimization (equation 1). DCGAN (deep convolutional generative adversarial network) specifically targeted the generation of images from random latent vectors (Radford et al., 2016). It borrowed from developments on convolutional neural networks to improve the sample quality and training stability on several datasets: using fully convolutional architectures (Springenberg et al., 2015), batch normalization (Ioffe & Szegedy, 2015), ReLU activation in the generator (Nair & Hinton, 2010), and Leaky ReLU activation in the discriminator (B. Xu et al., 2015). Its key advantage resides in its simplicity, and most GAN variants derive from this architecture. Among those variants, BigGAN represented a key milestone in the generation of large images (Brock et al., 2019). Its core contribution was to show that GANs benefit from scaling. First, BigGAN is two to four times deeper than previous GANs thanks to residual blocks (He et al., 2015). A residual block uses a shortcut connection in which the input to a stack of convolution layers is added to the output of that stack. It allows us to train deeper networks, improving sample quality. Second, each iteration during training uses a batch of data, and BigGAN uses a batch size height time larger than previous GANs. This is mainly possible thanks to new hardware developments, but it also improves sample quality. BigGAN also relies on combining other techniques proposed in previous studies, such as skip connections from the latent vector to multiple layers using conditional batch normalization (de Vries et al., 2017), orthogonal initialization of the convolution layers (Saxe et al., 2014), spectral normalization of the convolution layers (Miyato et al., 2018), and $R_1$ regularization of the discriminator (Mescheder et al., 2018). Compared to other architectures (e.g., Karras et al., 2018; Esser et al., 2021; Karras et al., 2021), BigGAN preserves some of the simplicity of DCGAN while achieving state-of-the-art results for GANs, although this comes with a high computational cost.

Both DCGAN and BigGAN were developed to generate 2D images. Extending those models to 3D is as straightforward as adding an extra dimension to the (transposed) convolution layers (Wu et al., 2016). This configuration has been used in other GAN models targeting subsurface applications (Laloy et al., 2018; Song et al., 2022), but it generates cubic images, i.e., all dimensions have the same size. Extracting cubic samples from our training realizations would have little value and, in practice, process-based realizations are often thinner than wide. Some models use a 4D latent tensor as input whose last three dimensions – the spatial dimensions – are adapted so that all (transposed) convolution layers affect each dimension in the same manner (Jo et al., 2020; Bhavsar et al., 2024). For instance, a latent vector with spatial dimensions of $8 \times 8 \times 1$ going through four transposed convolution layers will generate a sample of size $128 \times 128 \times 16$ cells. We propose a slightly different approach in which all dimensions start from the same size and stop growing as soon as the final sample size for a dimension is reached. For instance, a latent vector with spatial dimensions of $4 \times 4 \times 4$ going through five transposed convolution layers will generate a sample of size $128 \times 128 \times 16$ cells if the first two layers have a stride of $2 \times 2 \times 2$ and the last three of $2 \times 2 \times 1$. To preserve memory, the kernel size is also reduced to 1 along the dimensions that are not growing. For instance, the first two layers of the previous example have a kernel size of $4 \times 4 \times 4$ and the last three of $4 \times 4 \times 1$. This approach follows the scaling benefit highlighted by BigGAN in a parsimonious way: larger dimensions are allocated more depth and parameters to ensure a good sample quality along those dimensions.

## 2.3 Validation

Assessing GAN performance is harder than for other machine-learning approaches because of the minimax optimization: its objective function (equation 1) aims at finding a balance between generator and discriminator, so it tells us little about the quality and diversity of the generated samples. The two standard metrics used in computer science are the inception score (Salimans et al., 2016) and the Fréchet inception distance (Heusel et al., 2017). But those metrics both use a pre-trained deep-learning model for image classification, Inception v3 (Szegedy et al., 2015), which is limited to 2D images.

Another approach consists in comparing the generated samples with some validation samples using the sliced Wasserstein distance. The Wasserstein distance measures the optimal cost to transform a probability distribution into another, i.e., the amount of probability mass that needs to be moved times the distance over which it needs to be moved. While computing the Wasserstein distance in 1D is efficient, its cost gets prohibitive in higher dimensions. But we can approximate it by taking the average of the 1D Wasserstein distance from 1D distributions randomly extracted from the original nD distributions (Rabin et al., 2012). In the context of GANs, Karras et al. (2018) proposed to extract patches from images at multiple levels of a Laplacian pyramid and use the sliced Wasserstein distance on those patches to compare generated and validation samples. Thus, this approach assesses the reproduction of patterns at different level of details no matter their exact location in a sample. A similar strategy is already used in geological modeling (Tan et al., 2014), and the sliced Wasserstein distance has been used in multiple studies related to GANs and subsurface applications (e.g., Song et al., 2021; Zheng & Zhang, 2022; Hu et al., 2024). Here, we use the average distance to the validation set to assess performance during training. After training, we use multidimensional scaling to visualize the distance between some generated samples and the test set (Song et al., 2021). This allows us to assess not just sample quality but also diversity.

On top of the standard approach, we propose to explore whether geological concepts can be used to assess GAN performance. Indeed, process-based models let us simulate properties other than those of direct practical interest. FluvDepoSet includes the deposition time of the sediments, which the GAN has to generate in addition to the fraction of coarse sediments. Being in an undeformed environment, we expect the generated samples to honor the law of superposition: deposition time in a cell cannot be lower than that of the cell below. The fraction of cells in a sample that honors the law of superposition might be used for validation: initially, the GAN is expected to generate samples close to random, where this fraction should be around 0.5. As training progresses, it should converge to 1. The benefit of such metric comes from its interpretability – it directly relates to a core geological principle and is bounded between 0 and 1 – and its self-sufficiency – it does not require a validation set.

## 3 Results

All the models presented in this section were trained on two NVIDIA A100 80GB GPUs.

### 3.1 Architecture and hyper-parameter tuning

As its name implies, BigGAN is a massive network with a high computational cost. A more parsimonious model, as close to DCGAN as possible, would be much more valuable in practice. An ablation study can tell us which features of BigGAN are needed to generate fluvial deposits: starting from DCGAN, we progressively add some components of BigGAN and monitor their impact on performance (figure 1). All the models are trained over the same number of generator iterations, without early stopping. This aims to assess training stability. We also train three models for each configuration to assess the impact of randomness in the training process, i.e., the initialization of the neural networks and the selection of the validation set. Training more models would have given us a more robust picture, but computational cost limited what we could do.

A first observation is that DCGAN's training collapses completely for all models (figure 1, architecture 0). This architecture ends the discriminator with a sigmoid activation function, which is then used in a binary cross entropy loss function. Moving the sigmoid function inside the loss function – now called binary cross entropy with logits – takes advantage of the log-sum-exp trick to improve numerical stability. Combined with Leaky ReLU activation functions in the generator and changing the $\beta$ parameters of the Adam optimizer used for training (Kingma & Ba, 2015), it has a direct impact: while training is unstable and samples lack details, the non-stationarity of the fraction of coarse sediments and of the deposition time start to appear (figure 1, architecture 1). Adding regularization through spectral normalization on the other end leads to a less stable training with DCGAN's architecture (figure 1, architecture 2).

Introducing residual blocks leads to completely different results (figure 1, architecture 3): training remains stable for all three models and detailed deposits appear in the samples. This is supported by the sliced Wasserstein distance but also by the fraction of cells honoring the law of superposition, which tends toward 1 for all three models. Adding extra regularization on top of the spectral normalization has little to no impact on performance (figure 1, architecture 4). Using no regularization leads to poor performance (figure A.1, architecture 3b), but both spectral normalization and $R_1$ regularization perform equally well (figure 1, architecture 3, and figure A.1, architecture 4d).

The remaining components have little to no impact on training stability and samples' quality (figure 1, architectures 4 to 8). This includes doubling the residual blocks, so increasing the number of parameters (figure 1, architecture 5). Training with all those parameters remains stable, but it suggests that scaling has its limits for this specific training dataset. Compared to BigGAN, we did not add the orthogonal regularization and the components acting during sampling – most notably the truncation trick, which trades sample diversity for quality. A key issue we faced is the memory requirement of the full model (figure 1, architecture 8), even with two state-of-the-art GPUs with a large memory. Deep neural networks usually use a 32-bit floating point format; here, we used mixed precision training (Micikevicius et al., 2018), in which some operations use a 16-bit floating point format to lower the memory requirements and speed-up computation. BigGAN has two architectures, one using normal residual blocks and a more economical one using residual blocks with bottleneck, in which the size of the input is reduced before going into a stack of two convolution layers. To save memory and computation time, we used residual blocks with bottleneck as originally defined by He et al. (2015), i.e., the stack has a single convolution layer. We also applied $R_1$ regularization in a lazy manner, i.e., we only added it every 16 iterations (Karras et al., 2020). Brock et al. (2019) observed a degradation of performance with the original weight and solved it by reducing it. We found that keeping the original weight but applying it lazily was enough to stabilize training without hurting performance, with the extra benefit of reducing computation time. Even then, we limited our batch size to 64 instead of 2048 for BigGAN, and did not include the attention block in the generator and discriminator (Wang et al., 2018): its memory requirement in 3D was just too high no matter our attempts at reducing it.

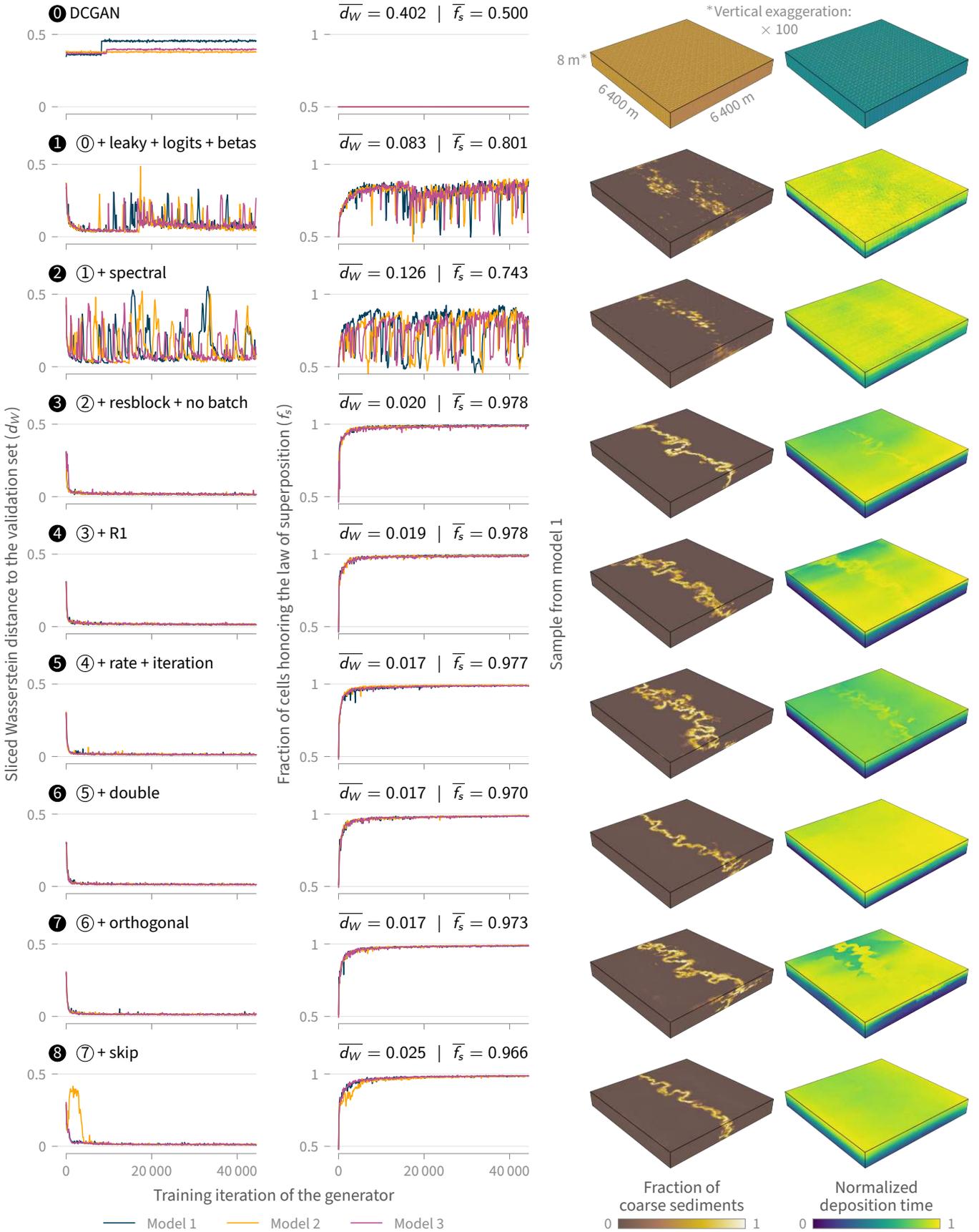

**Figure 1** Ablation study starting from DCGAN and progressively adding elements of BigGAN to assess their effect on training and sample quality. The samples come from the same latent vector. *leaky*: Leaky ReLU in the generator instead of ReLU; *logits*: binary cross entropy with logits as loss instead of binary cross entropy with a sigmoid function as last activation in the discriminator; *betas*: $\beta_1$ of 0 and $\beta_2$ of 0.99 instead of 0.5 and 0.999; *spectral*: spectral normalization in the generator and discriminator; *resblock*: residual blocks in the generator and discriminator instead of the convolutional layers; *no batch*: no batch normalization in the discriminator; *R1*: $R_1$ regularization; *rate*: learning rate of 0.00005 for the generator instead of 0.0002; *iteration*: two iterations of the discriminator per iteration of the generator; *double*: double residual blocks; *orthogonal*: orthogonal instead of normal initialization in the convolutional layers; *skip*: skip connections from the latent to the residual blocks.

These results highlight that GANs' training can be stable, even when trained on multiple properties from a process-based model. It is likely that the diversity of structures in the training samples is lower than that of the datasets used by the deep-learning community, which explains this stability. Indeed, BigGAN remained unstable when trained on ImageNet, and training had to be stopped early Brock et al. (2019). Interestingly, adding an extra dimension does not seem to impact performance: BigGAN was trained on 2D images up to $512 \times 512$ pixels, which is the same total number of cells than our $128 \times 128 \times 16$ 3D images. So developments meant to generate larger 2D images are directly transferable to 3D images. From a validation perspective, the fraction of cells that honor the law of superposition closely follows the variations of the sliced Wasserstein distance during training (figure 1, architectures 1 and 8). This is particularly valuable from a practical perspective, as mentioned in section 2.3: it confirms that geological principles can be used to monitor the performance of GANs.

### 3.2 Testing a parsimonious architecture

From a practical perspective, the more parsimonious the architecture, the better. In our ablation study, the most parsimonious architecture would be architectures 3 (figure 1) or 4d (figure A.1), i.e., a DCGAN with residual blocks, the binary cross entropy with logits, and one form of regularization. Combining the spectral normalization and the lazy $R_1$ regularization does not hurt performance nor computation time, but it might still help the GAN to generalize better to different training configurations. Thus, to keep our study tractable, we will only use architecture 4 (figure 1) from now on. We will only show results from model 1 in the main text; results for model 2 and 3 can be found in appendix. All the samples were the first generated based on a specific seed, without cherry picking.

We have several ways of assessing the performance of this architecture. The most straightforward one is a visual assessment. All three models are capable of generating a wide range of channel belts that are indistinguishable from samples extracted from the testing and training sets for the most part (figures 2 and 3 for model 1, B.1 and B.2 for model 2, B.4 and B.5 for model 3). The non-stationarity in the coarse fraction of sediments and in the deposition time is always properly captured. Few samples still show some implausible deposits, for instance some disconnected coarse deposits on the side of the model while the deposition time shows no erosion or incision there (figure 4). This further highlights the value of the deposition time to assess performance.

Of course, a visual assessment remains subjective and can only target a small number of samples. The validation of the models during training (figure 1, architecture 4) already gave us a more quantitative idea of the performance of the models. We can push the analysis further on the test set, which was completely unused during the training process. Generated samples overlap well with samples extracted from the realizations of the test set – two samples per realization, one at each extremity (figure 2 for model 1, B.1 for model 2, B.4 for model 3). This suggests that all models can capture the range of patterns visible in the test set, so they did not suffer from mode collapse. By showing several test samples and their closest generated samples according to the sliced Wasserstein distance, we can also judge the quality of the multidimensional scaling. Some generated samples end up quite far from their closest test samples, and we should refrain from over-interpreting this visualization. Aside from the test set, we can also visualize the distribution of the fraction of cells honoring the law of superposition for 10 000 generated samples. All three models have a vast majority of samples above 0.9 and a median above 0.99, highlighting the overall quality of the samples. Most violations to the law of superposition occur as individual cells in the channel belt, in larger areas along the borders of the samples, or along implausible sheet-like coarse deposits (figures B.1 and B.4 in particular).

Another possible issue with GANs is memorization of the training data. An easy way to assess it is to look for the closest training sample to a generated sample (Brock et al., 2019). Our default training configuration, where we randomly extract samples from realizations at each training iteration, makes this a cumbersome exercise. Instead, we retrain the three models of architecture 4 by extracting training samples deterministically. All three models train just as well as with random samples. Similarities can be observed between training and generated samples, but the samples remain clearly different. So the generator is not purely memorizing its training data.

A final test consists in assessing whether interpolating along the latent space leads to smooth transitions between different styles of deposits. Exploring the whole hundred-dimensional latent space is of course impossible, but we can already have an idea based on interpolating between four samples (figure 4 for model 1, B.2 for model 2, B.5 for model 3). All models can smoothly transition between different samples, although that transition can be non-linear. Ideally, we would like the transition to happen without new style appearing in the model (figure B.2). But this is not the case everywhere in the latent spaces, where some transitions between samples are not the most direct (figure 4). This suggests that at least part of the latent space is entangled, which might be an issue for downstream applications relying on optimizing a latent vector.

### 3.3 Sensitivity to the training dataset and latent size

After testing the performance of architecture 4, we can explore two key aspects for its practical use. The first one is the relationship between number of training samples and performance. Process-based models tend to have a high computational cost, especially in 3D, so having a generative model that can generalize from as few samples as possible is a great asset. Here we use again a training configuration with fixed samples extracted from the training realizations instead of random ones. Each training set consists of the first 190, 950, 1 900, 9 500, or 19 000 realizations of FluvDepoSet, while the validation set includes realizations 19 001 to 20 000. Considering that datasets used by the deep-learning community often contains several hundreds of thousands of images, these are very small to relatively small datasets. Surprisingly, even models with very few samples perform well (figure 5). The fraction of cells honoring the law of superposition clearly shows that training is not completely stable with less than 1 000 training samples. Still, the non-stationarity of the fraction of coarse sediments and of the deposition time are well captured, and details of the meanders starts to appear. At 1 900 training samples, the fraction of cells honoring the law of superposition shows again a small divergence after 30 000 iterations, but the quality of the samples is much better. Performance with 9 500 samples is just as good as with 19 000 samples. This suggests that a few thousand samples are enough to train models based on architecture 4. Yet, generalizing this observation to other datasets is difficult: datasets used by the deep-learning community for instance are larger because their samples capture a more diverse range of patterns.

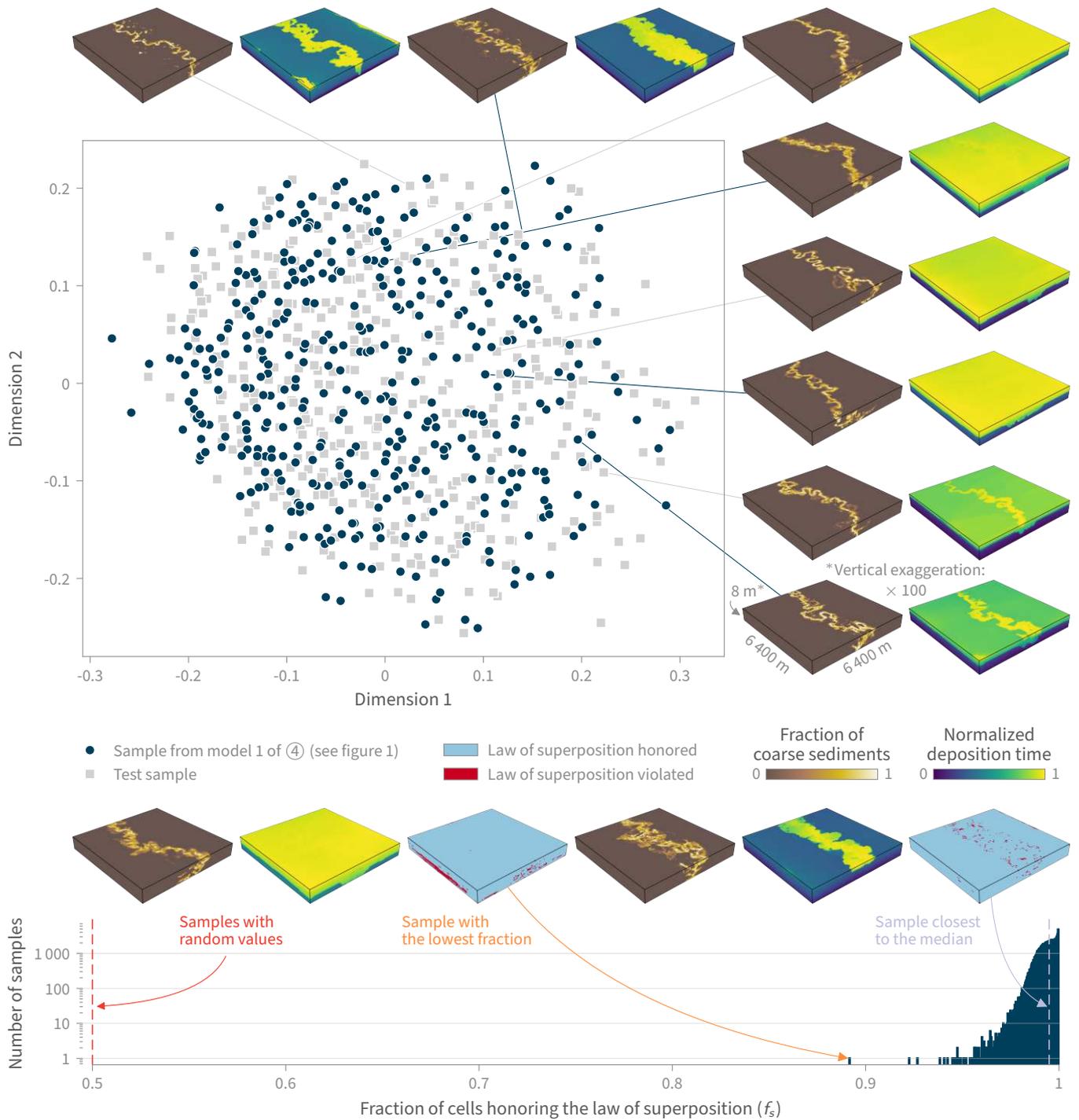

**Figure 2** Testing sample quality and diversity from model 1 of architecture 4 from figure 1 using multidimensional scaling to represent the sliced Wasserstein distances between 400 test samples and 400 samples from model 1 and the fraction of cells honoring the law of superposition in 10 000 samples from model 1.

The second key aspect to explore is the size of the latent space. From a training perspective, changing the size of the latent vector has little impact on the sliced Wasserstein distance and the fraction of cells honoring the law of superposition (figure 6). This suggests that the GAN is capable of organizing a latent space that captures the diversity of training samples, even based on a small latent vector. But this must come at a price, and a low-dimensional latent space is likely to be more entangled, which can create problems for downstream applications.

In addition, the generator's input is a 4D tensor, with the first dimension containing the latent vector, and the last three dimensions being the spatial dimensions with a size of 1 during training. While BigGAN uses a fully connected layer at the beginning of the generator, we have preserved in all our architectures the fully convolutional configuration of DCGAN. So we can increase the size of those spatial dimensions to generate larger samples (figure 6). Increasing the size of each spatial dimension by 1 increases the size of the samples by $32 \times 32 \times 4$ cells. Doing so essentially leads the GAN to extrapolate. This works relatively well in the directions where properties are stationary, so parallel to the channel belt and vertically for the coarse fraction of sediments, parallel and perpendicular to the channel belt for the deposition time. Extending the sample in the direction perpendicular to the channel belt leads to disconnected pieces of channel belts. Extending the sample vertically leads to a deposition time that clearly violates the law of superposition. This suggests that we can still generate larger samples when the training samples are stationary, which is the standard expectation in geological modeling. Even if we cannot control the size of those samples down to the cell, the extra cost of generating a slightly larger sample and truncate it is negligible. But a different strategy will be needed to extrapolate from non-stationary training samples, which is a likely case with training data from process-based models.

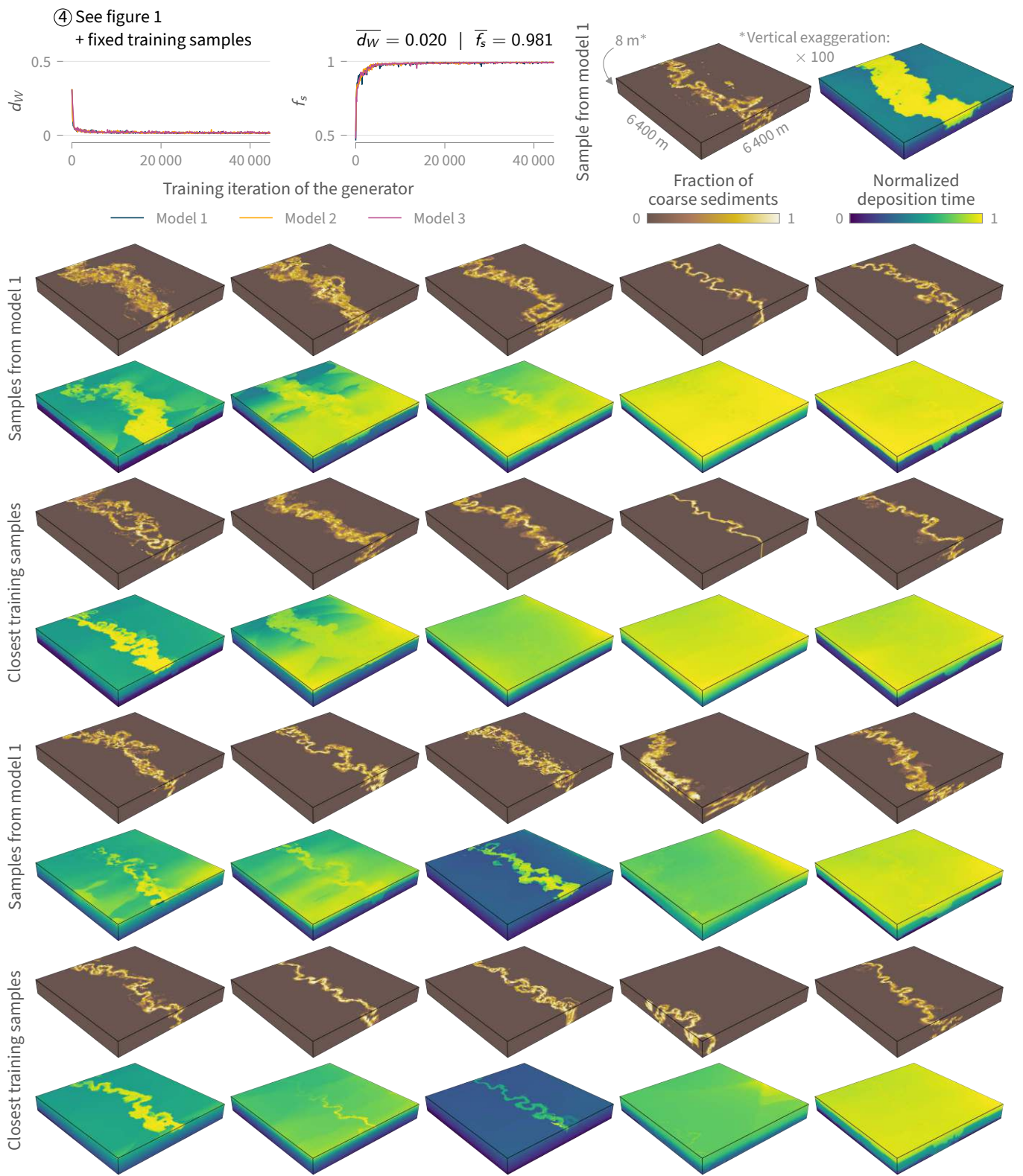

**Figure 3** Testing memorization of model 1 of architecture 4 from figure 1 using a training with fixed instead of random samples extracted from FluvDepoSet. $d_W$: sliced Wasserstein distance to the validation set; $f_s$: fraction of cells honoring the law of superposition.

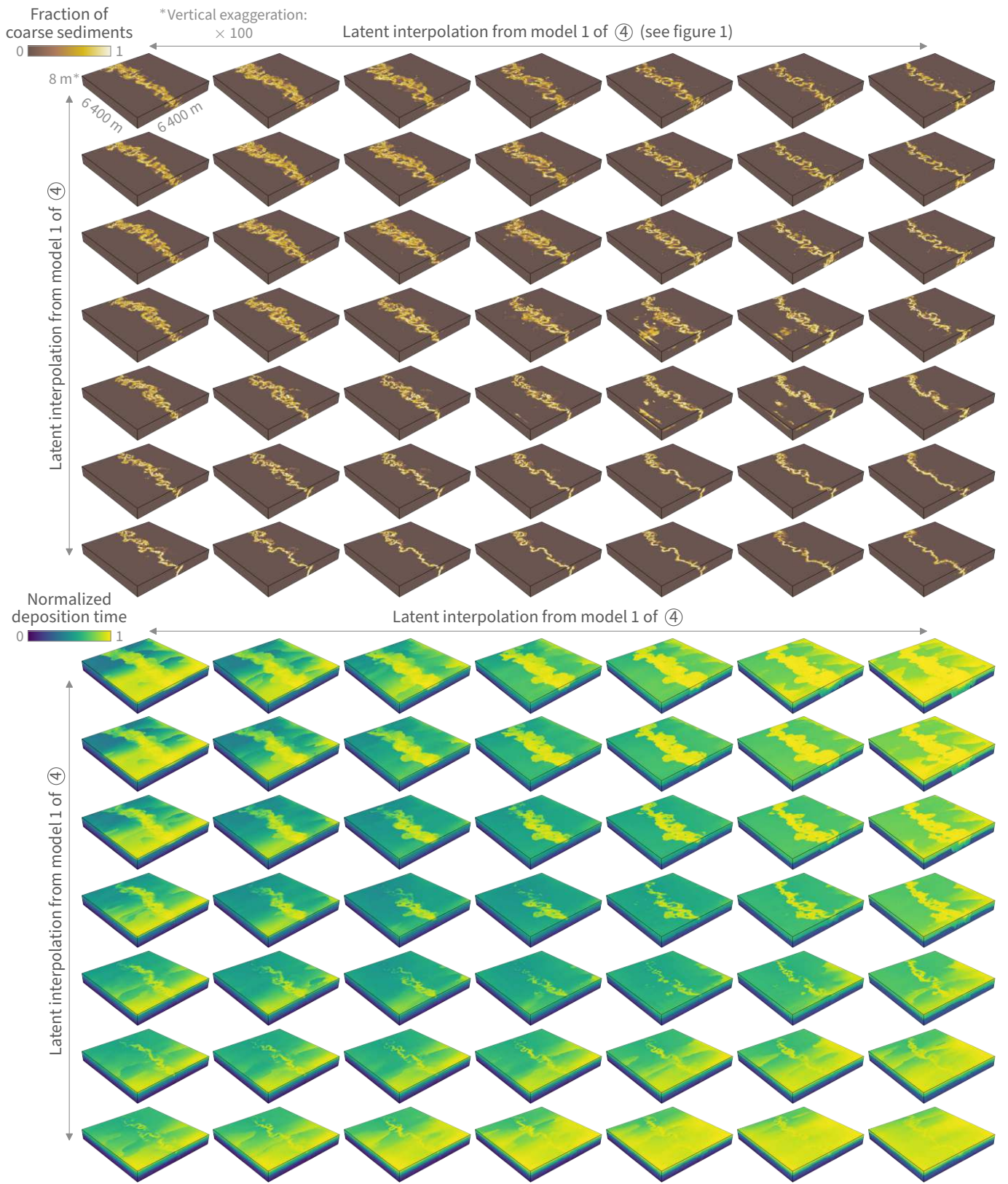

**Figure 4** Latent interpolation between four samples (at each corner) of model 1 of architecture 4 from figure 1.

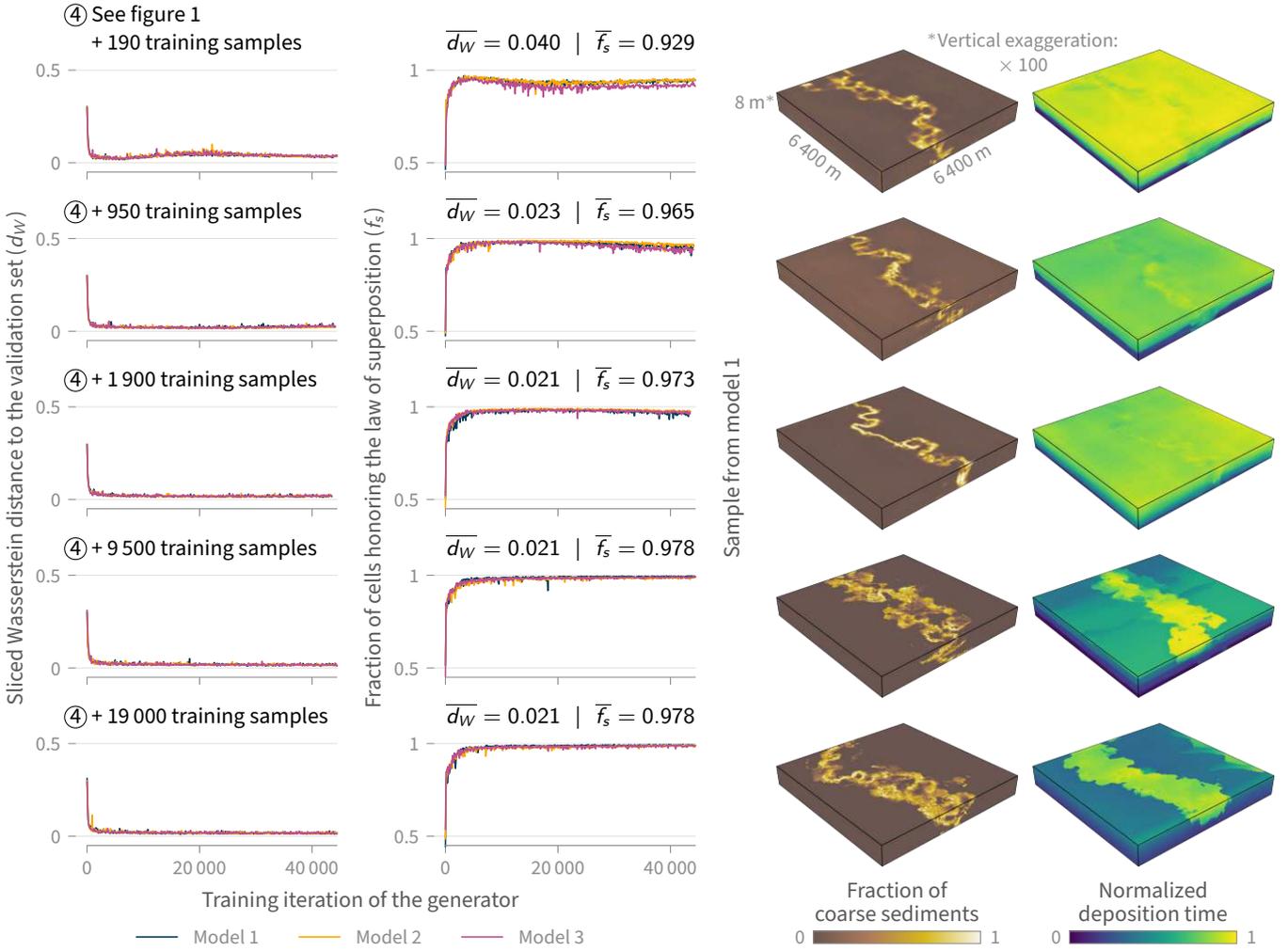

**Figure 5** Impact of the number of training samples on training and sample quality of model 1 of architecture 4 from figure 1. The samples come from the same latent vector.

### 3.4 Sensitivity to the training data

Now that we have explored the performance and robustness of architecture 4 trained on FluvDepoSet, a key question remains: how would it perform with other training datasets? Building a proper training dataset remains a big effort – simulating a realization from CHILD can easily take dozens of hours – and no other open dataset that we know of meets our requirements. Nevertheless, we can play with FluvDepoSet to explore other configurations (figure 7).

A first configuration is to generate samples where the channel belt is not necessarily centered. Channel belts' in FluvDepoSet realizations can migrate to the border of the domain (figure 3), but it remains rare. Here, we randomly extract from the training realizations samples of size $64 \times 64 \times 16$ so that each sample contains at least part of the channel belt. Training remains just as stable as with the original $128 \times 128 \times 16$ case, and performance as well. Having smaller samples might play a role in this, but this is encouraging regarding the robustness of GANs.

A second configuration is to mimic what was done in previous studies: generating a single discrete property, the facies (e.g., Laloy et al., 2018; Song et al., 2022; Bhavsar et al., 2024). To get facies, we computed the mean grain size using the fraction of coarse sediments with the values of coarse and fine grain diameter used to simulate the realization. We applied Folk (1954)'s classification to get three lithofacies: one with good aquifer properties (clayey sand & sand), one with poor aquifer properties (clay), and one intermediate (sandy clay). We then trained three models of architecture

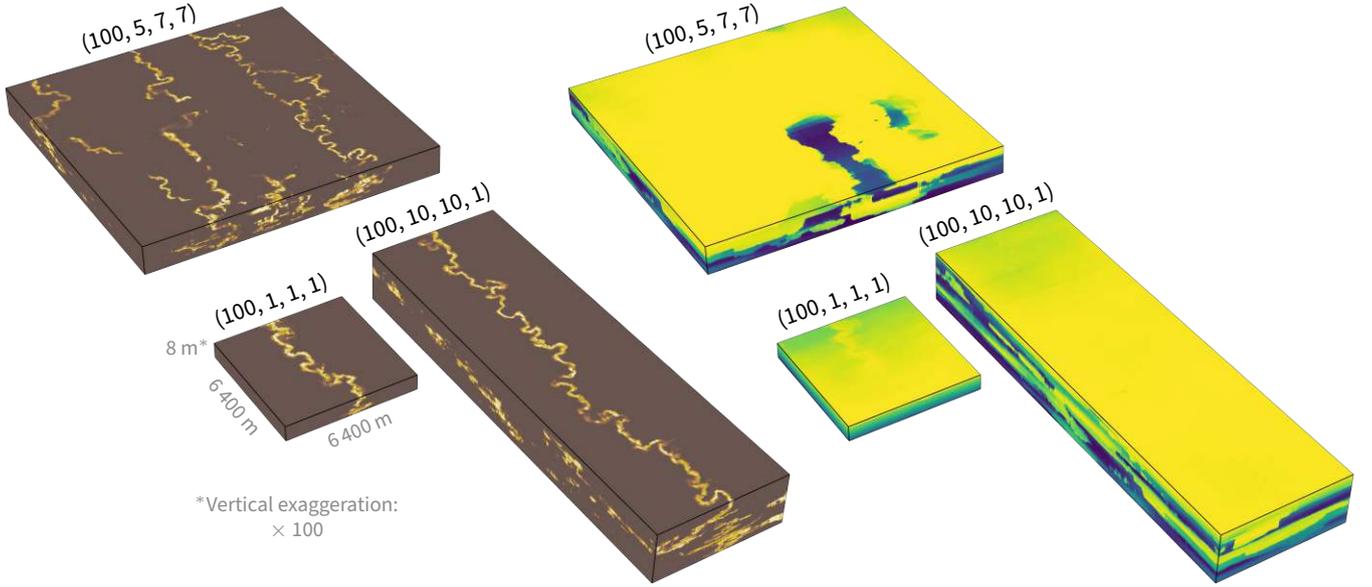
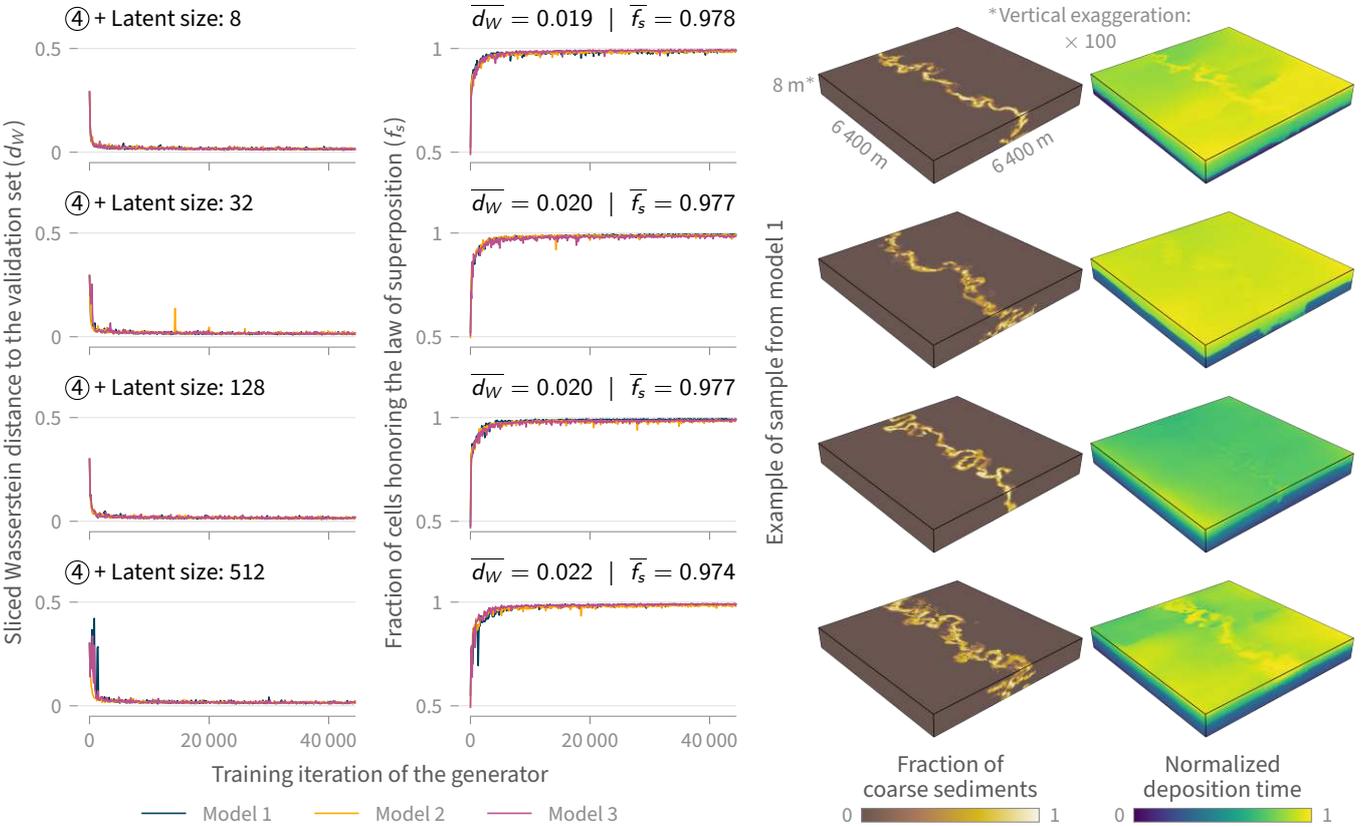

**Figure 6** Impact of the latent shape and size on sample quality. The samples with different latent sizes come from the same latent vector, which is truncated to generate smaller samples.

4 in the original $128 \times 128 \times 16$ case with a centered channel belt and in the $64 \times 64 \times 16$ case with a non-centered channel belt. In this setting, we do not generate the deposition time, so we cannot use the fraction of cells honoring the law of superposition for validation. The samples still show a good visual quality. We clearly see three distinct facies, although we did not change the last activation function of the generator to one more adapted to a discrete property or used indicator variables as done is other studies (e.g., Pan et al., 2021; Sun et al., 2023b; Bhavsar et al., 2024). The clayey sand & sand facies shows a similar continuity to the training samples and the sandy clay facies is indeed mostly in between the other two. The sliced Wasserstein distance confirm this assessment, with average values over training just slightly above the continuous case. Similar results ensue when training on the fraction of coarse sediments without the deposition time (figure A.2), suggesting two findings: first, generating the deposition time stabilizes training; second, generating discrete properties is as stable as generating continuous ones, highlighting again some robustness from architecture 4.

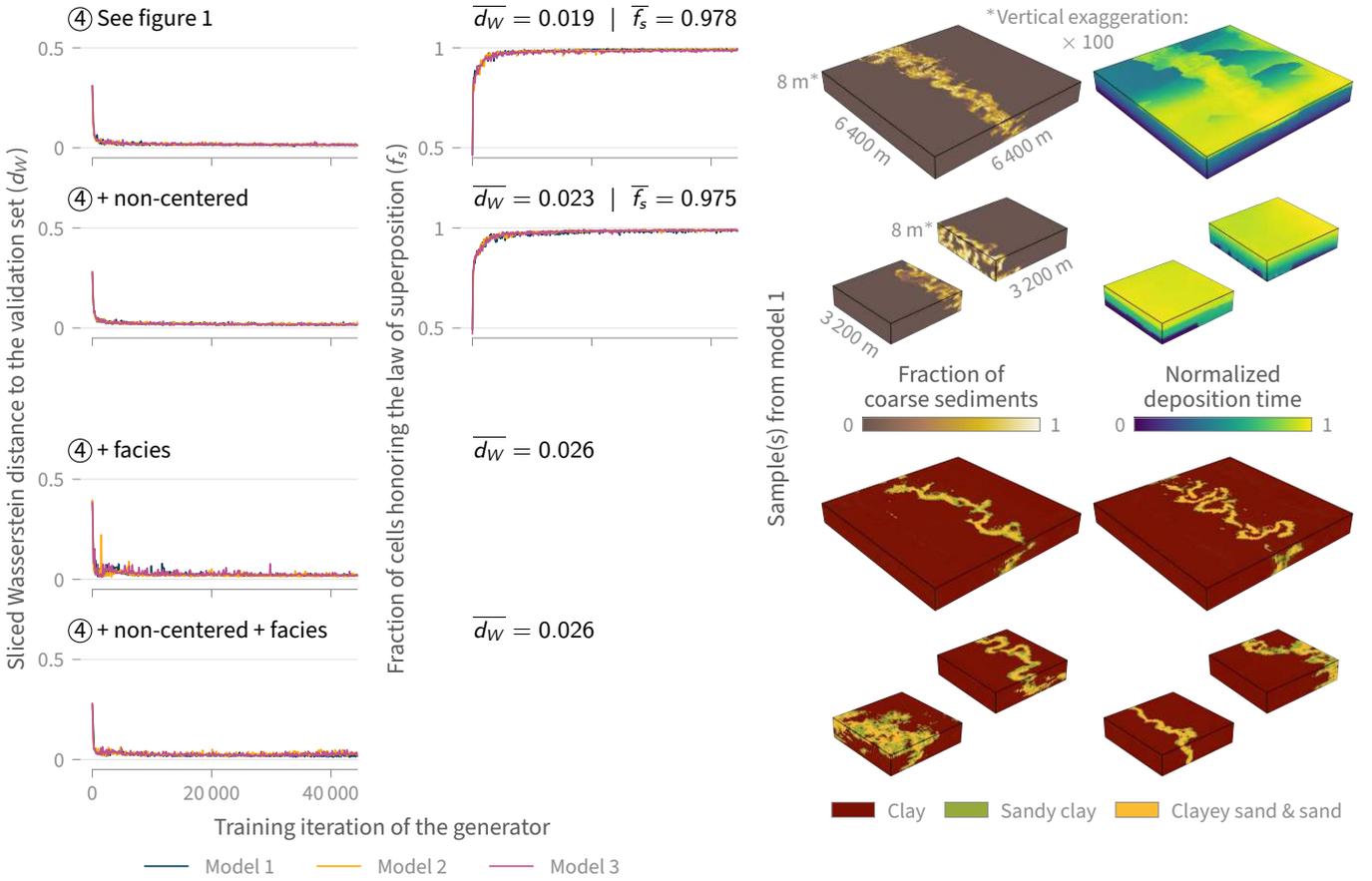

**Figure 7** Impact of the characteristics of the properties in the training samples on training and sample quality. Architecture 4 is the same as in figure 1. The samples come from the same latent vector. *non-centered*: smaller training samples in which the channel belt can be anywhere along the $x$ axis; *facies*: training samples with 3 facies based on the mean grain size instead of the fraction of coarse sediments; the deposition time is removed from those samples, so the law of superposition cannot be used for validation.

## 4 Discussion

Our work joins a series of previous studies suggesting that existing GAN architectures transfer well to subsurface applications (e.g., Laloy et al., 2018; Zheng & Zhang, 2022; Bhavsar et al., 2024): training remains stable and leads to appropriate sample quality and diversity for practical use. This should come as no surprise since the deep-learning community tests its models on large, diverse datasets. ImageNet for instance includes more than one million images divided into more than a thousand categories, from animals to airplanes (Russakovsky et al., 2015). The main difference resides in the dimensions of those images: 2D in the deep-learning community, 3D in subsurface applications. But those previous studies and our work shows that adding an extra dimension does not impede performance. This suggests that we can move from studies demonstrating the potential of GANs to generate subsurface properties to more in-depth studies assessing their practical value.

From this perspective, the philosophy behind generative modeling differs in the deep-learning community and in the subsurface community. The deep-learning community has mainly focused on developing huge foundation models built on humongous datasets to, ultimately, generate anything. The subsurface community, through the development of multiple-point simulation (e.g., Strebelle, 2002; Mariethoz et al., 2010; Hoffimann et al., 2017), has focused on small specialized models built on a case-by-case basis based on limited training data – what the machine-learning community would call one-shot or few-shot learning. In which direction can we take GANs for subsurface applications? Architecture 4 takes about 7 h to train over 150 epochs with a training dataset of 20 000 realizations. This time could be reduced by using a smaller training dataset and training over a smaller number of epochs since our validation metrics converge quickly (figure 5). Integrating developments for few-shot learning with GANs (Li et al., 2022) could further stabilize training with very few data and reduce its cost. While CHILD is computationally expensive, it remains a research software with room for optimization. Simpler process-based models could also simulate valuable realizations at a much lower cost (e.g., Howard & Knutson, 1984; Pyrcz et al., 2009). All in all, it should already be possible to generate some training data and train architecture 4 in less than 24 h, which remains reasonable compared to what is invested in downstream applications such as history matching. But further tests are needed to assess whether training GANs on a case-by-case basis is truly worthwhile. Architecture 4 has shown robustness to some variations in the training data (figure 7), which is encouraging in the perspective of building a foundation model. But a foundation model will only be useful if it targets a large variety of depositional environments, e.g., meandering, braided, deltaic, shelf, lake, and eolian deposits. A key issue then is the lack of investment in the development of process-based models. At the moment, we lack process-based models to simulate all the environments mentioned above at reservoir scale in an integrated framework, and too few of those models are open. Another key issue will be the difficulty of training GANs on a multimodal dataset with large samples to capture transitions between environments. While other components can still be added to stabilize training, even BigGAN is not fully stable with ImageNet (Brock et al., 2019), and the cost of training will be huge.

This raises the question of how to further improve performance. Even with architecture 4, the law of superposition remains partly violated, which suggests some room for improvement. While our hyper-parameter tuning was not as thorough as Brock et al. (2019)'s in the original BigGAN article, tests with different learning rates, $\beta$ parameters, or extra iterations of the discriminator did not improve performance (figures 1 and A.1). We also tested different initial architectures than DCGAN, and Wasserstein GAN (WGAN) in particular (Arjovsky et al., 2017; Gulrajani et al., 2017). WGAN is one of the few developments to make GAN more stable based on mathematical theory. In practice, it simply changes the loss function, making it usable with any architecture. Several studies have successfully used it in a subsurface context (e.g., Chan & Elsheikh, 2019; Song et al., 2022; Bhavsar et al., 2024). In our case however, we could not find a configuration where training was fully stable (figure A.3). This is consistent with Brock et al. (2019), who made the same observation when developing BigGAN. Of course, it does not mean that stable training is impossible with WGAN and FluvDepoSet, and a more thorough ablation study might solve this issue. But this is where training cost really becomes a problem: training a single model once would be perfectly doable with less performant GPUs, but training multiple models multiple times to account for different hyper-parameters becomes quickly prohibitive. On top of the hardware cost, there is an environmental cost linked to the energy needed to train such big models.

So should we keep developing GANs for subsurface applications? BigGAN remains part of GANs' state of the art several years after its release simply because the deep-learning community has moved on to a different type of deep generative models: diffusion models (e.g., Ho et al., 2020; Hoogeboom et al., 2021; Rombach et al., 2022). Diffusion models are more stable than GANs, with a more conventional training procedure less likely to collapse. So they better preserve sample diversity than GANs while generating high-quality samples. But this comes at a cost: generating samples with a diffusion model takes much longer than with a GAN. While few studies have explored their use in subsurface applications yet, the first results are encouraging (Di Federico & Durlofsky, 2024; Lee et al., 2025). So far, deep generative modeling has been dominated by GANs, diffusion models, and another type of models: variational autoencoder (VAEs) (e.g., Kingma & Welling, 2014; Higgins et al., 2017; van den Oord et al., 2017). And those models have complementary benefits and weaknesses, which Xiao et al. (2022) refers to as the generative learning trilemma:

- GANs can generate high-quality samples quickly but are prone to collapse.

- Diffusion models can generate high-quality and diverse samples but are slow.

- VAEs can generate diverse samples quickly but at low quality.

We believe that exploring the use of all types of deep generative models in subsurface applications is valuable. But this only stands if we move away from articles focusing on advertising the success of new approaches to articles focusing on exploring the limitations of those approaches. Where models break is often more insightful than where they work.

No matter where we focus our efforts, we must keep in mind geological principles and processes. It is hard to predict how deep generative modeling will evolve but, despite their successes, current models remain statistical models on steroids. This becomes obvious when asking a GAN to extrapolate (figure 6): all it does is reproducing the final geometry of the deposits, without any real understanding of the geological principles behind that geometry. This limitation is only obvious because we had deposition time in our training data, and not just the usual properties of geological models such facies, porosity, or permeability. On top of investing in simulating plausible synthetic training data, we need to invest in the development of deep generative models that properly extrapolate, including in non-stationary settings. Current approaches relies on expanding existing patches (Abdellatif & Elsheikh, 2023; Pan et al., 2023; Sun et al., 2023a), but this might be where specific architectural developments are most needed. Those developments should not be limited to the deep-learning perspective. Here, we showed that the law of superposition can support validation, and we strongly believe that leveraging geological principles in any way, shape, or form will only improve subsurface modeling.

## 5 Conclusions

While standard GAN architectures like DCGAN struggle to generate fluvial deposits, architectures aiming at generating larger images like BigGAN can do so in a stable way, including in 3D. Not all components of BigGAN are required to reach that stage: adding residual blocks, the binary cross entropy with logits, and spectral normalization or lazy $R_1$ regularization to DCGAN is enough to generate samples of quality and diversity comparable to the training dataset FluvDepoSet. A similar quality and diversity of generated samples can be obtained with continuous and discrete properties alike, with clear channel belts and detailed geometries linked to deposition along meanders. Contrary to standard generative approaches in geological modeling, a BigGAN-like model can capture the non-stationarity of the generated properties without any feature engineering. Generating the deposition time in the samples further stabilizes training and allows us to use the law of superposition for validation. Monitoring the fraction of cells honoring that law during training leads to the same observations as a metric capturing the differences in patterns with a validation set. The deposition time also allows us to have a better idea of the geological plausibility of the samples. This is useful not just to check individual samples, but also to assess whether the GAN is actually achieving what we think it achieves.

Our results suggest that combining GANs with process-based models for training data simulation is a viable approach for geological modeling. Yet, further studies are needed to test whether training GANs on a case-by-case basis is worthwhile, and whether training a foundation model to generate sedimentary deposits is feasible. Further comparison with other deep generative models such as diffusion models or VAEs is also needed. This comparison should not just focus on sample quality, but should also include sample diversity as well as training and prediction costs. Whichever model we use, investing in the simulation of plausible synthetic training data is an absolute necessity. Moving away from a pure deep-learning perspective and integrating geological principles into our models will be key to improve our prediction capabilities. Indeed, we are not interested in training a GAN for the sake of it, but to predict the physical properties of the subsurface. So conditioning to subsurface data is essential, an aspect that we will explore in the second part of this article.

## Data and software availability

Our entire study is openly available, including the scripts to reproduce results and figures, the results themselves, the pretrained models (Rongier & Peeters, 2025b), and the training data (Rongier & Peeters, 2021). Those scripts uses the open-source Python package voxgan (Rongier, 2021), which is built upon PyTorch (Ansel et al., 2024). The figures were made using the open-source Python packages matplotlib (Hunter, 2007) and PyVista (Sullivan & Kaszynski, 2019).

## Acknowledgments

This research was partly funded by the Deep Earth Imaging Future Science Platform, CSIRO. We would like to thank CSIRO's High Performance Computing team and the GSE Computational Team for their support.

## References

Abdellatif, A., & Elsheikh, A. H. (2023). *Generating Infinite-Resolution Texture using GANs with Patch-by-Patch Paradigm*. arXiv: 2309.02340 [cs, eess]. https://doi.org/10.48550/arXiv.2309.02340 (cited on page 13).

Ansel, J., Yang, E., He, H., Gimelshein, N., Jain, A., Voznesensky, M., Bao, B., Bell, P., Berard, D., Burovski, E., Chauhan, G., Chourdia, A., Constable, W., Desmaison, A., DeVito, Z., Ellison, E., Feng, W., Gong, J., Gschwind, M., … Chintala, S. (2024). PyTorch 2: Faster Machine Learning Through Dynamic Python Bytecode Transformation and Graph Compilation. *29th ACM International Conference on Architectural Support for Programming Languages and Operating Systems*, *2*, 929–947. https://doi.org/10.1145/3620665.3640366 (cited on page 14).

Arjovsky, M., Chintala, S., & Bottou, L. (2017). Wasserstein Generative Adversarial Networks. *34th International Conference on Machine Learning*, 214–223. https://doi.org/10.48550/arXiv.1701.07875 (cited on page 13).

Bengio, Y., Courville, A., & Vincent, P. (2013). Representation Learning: A Review and New Perspectives. *IEEE Transactions on Pattern Analysis and Machine Intelligence*, *35*(8), 1798–1828. https://doi.org/10.1109/TPAMI.2013.50 (cited on page 2).

Bhavsar, F., Desassis, N., Ors, F., & Romary, T. (2024). A stable deep adversarial learning approach for geological facies generation. *Computers & Geosciences*, *190*, 105638. https://doi.org/10.1016/j.cageo.2024.105638 (cited on pages 3, 10–13).

Bridge, J. S. (2007). *Rivers and floodplains: Forms, processes, and sedimentary record* (3rd ed.). Blackwell (cited on page 1).

Brock, A., Donahue, J., & Simonyan, K. (2019). Large Scale GAN Training for High Fidelity Natural Image Synthesis. *7th International Conference on Learning Representations*. https://doi.org/10.48550/arXiv.1809.11096 (cited on pages 2–4, 6, 13).

Chan, S., & Elsheikh, A. H. (2019). Parametric generation of conditional geological realizations using generative neural networks. *Computational Geosciences*, *23*(5), 925–952. https://doi.org/10.1007/s10596-019-09850-7 (cited on page 13).

Clevis, Q., De Boer, P. L., & Nijman, W. (2004). Differentiating the effect of episodic tectonism and eustatic sea-level fluctuations in foreland basins filled by alluvial fans and axial deltaic systems: Insights from a three-dimensional stratigraphic forward model. *Sedimentology*, *51*(4), 809–835. https://doi.org/10.1111/j.1365-3091.2004.00652.x (cited on page 1).

de Vries, H., Strub, F., Mary, J., Larochelle, H., Pietquin, O., & Courville, A. (2017). Modulating early visual processing by language. *31st International Conference on Neural Information Processing Systems*, 6597–6607. https://doi.org/10.48550/arXiv.1707.00683 (cited on page 3).

Deutsch, C. V., & Journel, A. G. (1992). *GSLIB: Geostatistical Software Library and User's Guide*. Oxford University Press (cited on page 1).

Deutsch, C. V., & Wang, L. (1996). Hierarchical object-based stochastic modeling of fluvial reservoirs. *Mathematical Geology*, *28*(7), 857–880. https://doi.org/10.1007/BF02066005 (cited on page 1).

Di Federico, G., & Durlofsky, L. J. (2024). Latent diffusion models for parameterization of facies-based geomodels and their use in data assimilation. *Computers & Geosciences*, 105755. https://doi.org/10.1016/j.cageo.2024.105755 (cited on page 13).

Donselaar, M. E., Bhatt, A. G., & Ghosh, A. K. (2017). On the relation between fluvio-deltaic flood basin geomorphology and the wide-spread occurrence of arsenic pollution in shallow aquifers. *Science of The Total Environment*, *574*, 901–913. https://doi.org/10.1016/j.scitotenv.2016.09.074 (cited on page 1).

Enemark, T., Peeters, L. J. M., Mallants, D., & Batelaan, O. (2019). Hydrogeological conceptual model building and testing: A review. *Journal of Hydrology*, *569*, 310–329. https://doi.org/10.1016/j.jhydrol.2018.12.007 (cited on page 1).

Esser, P., Rombach, R., & Ommer, B. (2021). Taming Transformers for High-Resolution Image Synthesis. *2021 IEEE/CVF Conference on Computer Vision and Pattern Recognition (CVPR)*, 12868–12878. https://doi.org/10.1109/CVPR46437.2021.01268 (cited on page 3).

Folk, R. L. (1954). The Distinction between Grain Size and Mineral Composition in Sedimentary-Rock Nomenclature. *The Journal of Geology*, *62*(4), 344–359. https://doi.org/10.1086/626171 (cited on page 10).

Gómez-Hernández, J. J., & Wen, X.-H. (1998). To be or not to be multi-Gaussian? A reflection on stochastic hydrogeology. *Advances in Water Resources*, *21*(1), 47–61. https://doi.org/10.1016/S0309-1708(96)00031-0 (cited on page 1).

Goodfellow, I. J., Pouget-Abadie, J., Mirza, M., Xu, B., Warde-Farley, D., Ozair, S., Courville, A., & Bengio, Y. (2014). Generative adversarial nets. *28th International Conference on Neural Information Processing Systems*, *2*, 2672–2680 (cited on pages 2, 3).

Granjeon, D., & Joseph, P. (1999). Concepts and Applications of a 3-D Multiple Lithology, Diffusive Model in Stratigraphic Modeling. *Special Publications of SEPM*, *62*, 197–210. https://doi.org/10.2110/pec.99.62.0197 (cited on page 1).

Guardiano, F. B., & Srivastava, R. M. (1993). Multivariate Geostatistics: Beyond Bivariate Moments. In A. Soares (Ed.), *Geostatistics Tróia '92* (pp. 133–144, Vol. 1). Springer Netherlands. https://doi.org/10.1007/978-94-011-1739-5_12 (cited on page 1).

Gulrajani, I., Ahmed, F., Arjovsky, M., Dumoulin, V., & Courville, A. (2017). Improved training of Wasserstein GANs. *31st International Conference on Neural Information Processing Systems*,

5769–5779. https://doi.org/10.48550/arXiv.1704.00028 (cited on page 13).

He, K., Zhang, X., Ren, S., & Sun, J. (2015). *Deep Residual Learning for Image Recognition*. arXiv: 1512.03385 [cs]. https://doi.org/10.48550/arXiv.1512.03385 (cited on pages 3, 4).

Heusel, M., Ramsauer, H., Unterthiner, T., Nessler, B., & Hochreiter, S. (2017). GANs trained by a two time-scale update rule converge to a local nash equilibrium. *31st International Conference on Neural Information Processing Systems*, 6629–6640. https://doi.org/10.48550/arXiv.1706.08500 (cited on page 3).

Higgins, I., Matthey, L., Pal, A., Burgess, C., Glorot, X., Botvinick, M., Mohamed, S., & Lerchner, A. (2017). Beta-VAE: Learning Basic Visual Concepts with a Constrained Variational Framework. *5th International Conference on Learning Representations*. Retrieved 2025, from https://openreview.net/forum?id=Sy2fzU9gl (cited on page 13).

Ho, J., Jain, A., & Abbeel, P. (2020). Denoising diffusion probabilistic models. *34th International Conference on Neural Information Processing Systems*, 6840–6851. https://doi.org/10.48550/arXiv.2006.11239 (cited on page 13).

Hoffimann, J., Scheidt, C., Barfod, A., & Caers, J. (2017). Stochastic simulation by image quilting of process-based geological models. *Computers & Geosciences*, *106*, 18–32. https://doi.org/10.1016/j.cageo.2017.05.012 (cited on page 12).

Hoogeboom, E., Gritsenko, A. A., Bastings, J., Poole, B., Berg, R. van den, & Salimans, T. (2021). Autoregressive Diffusion Models. *9th International Conference on Learning Representations*. https://doi.org/10.48550/arXiv.2110.02037 (cited on page 13).

Howard, A. D., & Knutson, T. R. (1984). Sufficient conditions for river meandering: A simulation approach. *Water Resources Research*, *20*(11), 1659–1667. https://doi.org/10.1029/WR020i011p01659 (cited on page 12).

Hu, X., Song, S., Hou, J., Yin, Y., Hou, M., & Azevedo, L. (2024). Stochastic Modeling of Thin Mud Drapes Inside Point Bar Reservoirs With ALLUVSIM-GANSim. *Water Resources Research*, *60*(6), e2023WR035989. https://doi.org/10.1029/2023WR035989 (cited on page 4).

Hunter, J. D. (2007). Matplotlib: A 2D Graphics Environment. *Computing in Science & Engineering*, *9*(3), 90–95. https://doi.org/10.1109/MCSE.2007.55 (cited on page 14).

Ioffe, S., & Szegedy, C. (2015). Batch normalization: Accelerating deep network training by reducing internal covariate shift. *32nd International Conference on International Conference on Machine Learning*, *37*, 448–456. https://doi.org/10.48550/arXiv.1502.03167 (cited on page 3).

Jo, H., Santos, J. E., & Pyrcz, M. J. (2020). Conditioning well data to rule-based lobe model by machine learning with a generative adversarial network. *Energy Exploration & Exploitation*, *38*(6), 2558–2578. https://doi.org/10.1177/0144598720937524 (cited on page 3).

Journel, A. G. (2005). Beyond Covariance: The Advent of Multiple-Point Geostatistics. In O. Leuangthong & C. V. Deutsch (Eds.), *Geostatistics Banff 2004* (pp. 225–233). Springer Netherlands. https://doi.org/10.1007/978-1-4020-3610-1_23 (cited on page 1).

Karras, T., Aila, T., Laine, S., & Lehtinen, J. (2018). Progressive Growing of GANs for Improved Quality, Stability, and Variation. *6th International Conference on Learning Representations*. https://doi.org/10.48550/arXiv.1710.10196 (cited on pages 2–4).

Karras, T., Aittala, M., Laine, S., Härkönen, E., Hellsten, J., Lehtinen, J., & Aila, T. (2021). Alias-free generative adversarial networks. *35th International Conference on Neural Information Processing Systems*, 852–863. https://doi.org/10.48550/arXiv.2106.12423 (cited on page 3).

Karras, T., Laine, S., Aittala, M., Hellsten, J., Lehtinen, J., & Aila, T. (2020). Analyzing and Improving the Image Quality of StyleGAN. *2020 IEEE/CVF Conference on Computer Vision and Pattern Recognition (CVPR)*, 8107–8116. https://doi.org/10.1109/CVPR42600.2020.00813 (cited on page 4).

Kingma, D. P., & Ba, J. (2015). Adam: A Method for Stochastic Optimization. *3rd International Conference for Learning Representations*. https://doi.org/10.48550/arXiv.1412.6980 (cited on page 4).

Kingma, D. P., & Welling, M. (2014). Auto-Encoding Variational Bayes. *2nd International Conference on Learning Representations*. https://doi.org/10.48550/arXiv.1312.6114 (cited on page 13).

Laloy, E., Hérault, R., Jacques, D., & Linde, N. (2018). Training-Image Based Geostatistical Inversion Using a Spatial Generative Adversarial Neural Network. *Water Resources Research*, *54*(1), 381–406. https://doi.org/10.1002/2017WR022148 (cited on pages 3, 10, 12).

Lee, D., Ovanger, O., Eidsvik, J., Aune, E., Skauvold, J., & Hauge, R. (2025). Latent diffusion model for conditional reservoir facies generation. *Computers & Geosciences*, *194*, 105750. https://doi.org/10.1016/j.cageo.2024.105750 (cited on page 13).

Li, Z., Xia, B., Zhang, J., Wang, C., & Li, B. (2022). *A Comprehensive Survey on Data-Efficient GANs in Image Generation*. arXiv: 2204.08329 [cs]. https://doi.org/10.48550/arXiv.2204.08329 (cited on page 12).

Mariethoz, G., Renard, P., & Straubhaar, J. (2010). The Direct Sampling method to perform multiple-point geostatistical simulations. *Water Resources Research*, *46*(11). https://doi.org/10.1029/2008WR007621 (cited on page 12).

Mescheder, L., Geiger, A., & Nowozin, S. (2018). Which Training Methods for GANs do actually Converge? *35th International Conference on Machine Learning*, 3481–3490. https://doi.org/10.48550/arXiv.1801.04406 (cited on page 3).

Micikevicius, P., Narang, S., Alben, J., Diamos, G., Elsen, E., Garcia, D., Ginsburg, B., Houston, M., Kuchaiev, O., Venkatesh, G., & Wu, H. (2018). Mixed Precision Training. *6th International Conference on Learning Representations*. https://doi.org/10.48550/arXiv.1710.03740 (cited on page 4).

Miyato, T., Kataoka, T., Koyama, M., & Yoshida, Y. (2018). Spectral Normalization for Generative Adversarial Networks. *6th International Conference on Learning Representations*. https://doi.org/10.48550/arXiv.1802.05957 (cited on page 3).

Morris, R. C., & Ramanaidou, E. R. (2007). Genesis of the channel iron deposits (CID) of the Pilbara region, Western Australia. *Australian Journal of Earth Sciences*, *54*(5), 733–756. https://doi.org/10.1080/08120090701305251 (cited on page 1).

Nair, V., & Hinton, G. E. (2010). Rectified linear units improve restricted Boltzmann machines. *27th International Conference on International Conference on Machine Learning*, 807–814 (cited on page 3).

Pan, W., Chen, J., Mohamed, S., Jo, H., Santos, J. E., & Pyrcz, M. J. (2023). Efficient Subsurface Modeling with Sequential Patch Generative Adversarial Neural Networks. *SPE Annual Technical Conference and Exhibition*. https://doi.org/10.2118/214985-MS (cited on page 13).

Pan, W., Torres-Verdín, C., & Pyrcz, M. J. (2021). Stochastic Pix2pix: A New Machine Learning Method for Geophysical and Well Conditioning of Rule-Based Channel Reservoir Models. *Natural Resources Research*, *30*(2), 1319–1345. https://doi.org/10.1007/s11053-020-09778-1 (cited on page 11).

Pyrcz, M. J., Boisvert, J. B., & Deutsch, C. V. (2009). ALLUVSIM: A program for event-based stochastic modeling of fluvial depositional systems. *Computers & Geosciences*, *35*(8), 1671–1685. https://doi.org/10.1016/j.cageo.2008.09.012 (cited on page 12).

Rabin, J., Peyré, G., Delon, J., & Bernot, M. (2012). Wasserstein Barycenter and Its Application to Texture Mixing. In A. M. Bruckstein, B. M. ter Haar Romeny, A. M. Bronstein, & M. M. Bronstein (Eds.), *Scale Space and Variational Methods in Computer Vision* (pp. 435–446). Springer. https://doi.org/10.1007/978-3-642-24785-9_37 (cited on page 4).

Radford, A., Metz, L., & Chintala, S. (2016). Unsupervised Representation Learning with Deep Convolutional Generative Adversarial Networks. *4th International Conference on Learning Representations*. https://doi.org/10.48550/arXiv.1511.06434 (cited on page 3).

Rasmussen, C. E., & Williams, C. K. I. (2006). *Gaussian processes for machine learning*. MIT Press. https://doi.org/10.7551/mitpress/3206.001.0001 (cited on page 1).

Rombach, R., Blattmann, A., Lorenz, D., Esser, P., & Ommer, B. (2022). High-Resolution Image Synthesis with Latent Diffusion Models. *2022 IEEE/CVF Conference on Computer Vision and Pattern Recognition (CVPR)*, 10674–10685. https://doi.org/10.1109/CVPR52688.2022.01042 (cited on page 13).

Rongier, G. (2021). *Voxgan*. https://doi.org/10.25919/CDGF-CW44 (cited on page 14).

Rongier, G., & Peeters, L. (2021). *FluvDepoSet*. CSIRO. https://doi.org/10.25919/4FYQ-Q291 (cited on pages 2, 14).

Rongier, G., & Peeters, L. (2025a). *FluvDepoSet: A dataset of synthetic 3D models of fluvial deposits*. EarthArXiv: 10507. https://doi.org/10.31223/X5HX8D (cited on page 2).

Rongier, G., & Peeters, L. (2025b). *FluvGAN: Python scripts to test generating and inverting fluvial deposits using GANs* (Version 1). https://doi.org/10.4121/3469C879-F443-4FCF-83A2-B6DF56E96714.V1 (cited on pages 2, 14).

Russakovsky, O., Deng, J., Su, H., Krause, J., Satheesh, S., Ma, S., Huang, Z., Karpathy, A., Khosla, A., Bernstein, M., Berg, A. C., & Fei-Fei, L. (2015). ImageNet Large Scale Visual Recognition Challenge. *International Journal of Computer Vision*, *115*(3), 211–252. https://doi.org/10.1007/s11263-015-0816-y (cited on page 12).

Salimans, T., Goodfellow, I., Zaremba, W., Cheung, V., Radford, A., & Chen, X. (2016). Improved techniques for training GANs. *30th International Conference on Neural Information Processing Systems*, 2234–2242. https://doi.org/10.48550/arXiv.1606.03498 (cited on page 3).

Saxe, A., McClelland, J. L., & Ganguli, S. (2014). Exact solutions to the nonlinear dynamics of learning in deep linear neural networks. *2nd International Conference on Learning Representations*. https://doi.org/10.48550/arXiv.1312.6120 (cited on page 3).

Song, S., Mukerji, T., & Hou, J. (2021). Geological Facies modeling based on progressive growing of generative adversarial networks (GANs). *Computational Geosciences*, *25*(3), 1251–1273. https://doi.org/10.1007/s10596-021-10059-w (cited on page 4).

Song, S., Mukerji, T., Hou, J., Zhang, D., & Lyu, X. (2022). GANSim-3D for Conditional Geomodeling: Theory and Field Application. *Water Resources Research*, *58*(7), e2021WR031865. https://doi.org/10.1029/2021WR031865 (cited on pages 3, 10, 13).

Springenberg, J. T., Dosovitskiy, A., Brox, T., & Riedmiller, M. (2015). Striving for Simplicity: The All Convolutional Net. *3rd International Conference on Learning Representations*. https://doi.org/10.48550/arXiv.1412.6806 (cited on page 3).

Strebelle, S. (2002). Conditional Simulation of Complex Geological Structures Using Multiple-Point Statistics. *Mathematical Geology*, *34*(1), 1–21. https://doi.org/10.1023/A:1014009426274 (cited on pages 1, 12).

Sullivan, C., & Kaszynski, A. (2019). PyVista: 3D plotting and mesh analysis through a streamlined interface for the Visualization Toolkit (VTK). *Journal of Open Source Software*, *4*(37), 1450. https://doi.org/10.21105/joss.01450 (cited on page 14).

Sun, C., Demyanov, V., & Arnold, D. (2023a). A conditional GAN-based approach to build 3D facies models sequentially upwards. *Computers & Geosciences*, *181*, 105460. https://doi.org/10.1016/j.cageo.2023.105460 (cited on page 13).

Sun, C., Demyanov, V., & Arnold, D. (2023b). Geological realism in Fluvial facies modelling with GAN under variable depositional conditions. *Computational Geosciences*, *27*(2), 203–221. https://doi.org/10.1007/s10596-023-10190-w (cited on page 11).

Szegedy, C., Liu, W., Jia, Y., Sermanet, P., Reed, S., Anguelov, D., Erhan, D., Vanhoucke, V., & Rabinovich, A. (2015). Going deeper with convolutions. *2015 IEEE Conference on Computer Vision and Pattern Recognition (CVPR)*, 1–9. https://doi.org/10.1109/CVPR.2015.7298594 (cited on page 3).

Tan, X., Tahmasebi, P., & Caers, J. (2014). Comparing Training-Image Based Algorithms Using an Analysis of Distance. *Mathematical Geosciences*, *46*(2), 149–169. https://doi.org/10.1007/s11004-013-9482-1 (cited on page 4).

Tetzlaff, D. M., & Harbaugh, J. W. (1989). *Simulating Clastic Sedimentation* (1st ed., Vol. 1110). Springer. Retrieved 2024, from https://link.springer.com/book/9781475706949 (cited on page 1).

Tomczak, J. M. (2022). *Deep Generative Modeling*. Springer International Publishing. https://doi.org/10.1007/978-3-030-93158-2 (cited on page 2).

Tucker, G. E., Lancaster, S. T., Gasparini, N. M., & Bras, R. L. (2001). The Channel-Hillslope Integrated Landscape Development Model (CHILD). In R. S. Harmon & W. W. Doe (Eds.), *Landscape Erosion and Evolution Modeling* (pp. 349–388). Springer US. https://doi.org/10.1007/978-1-4615-0575-4_12 (cited on page 2).

Tucker, G. E., Lancaster, S. T., Gasparini, N. M., Bras, R. L., & Rybarczyk, S. M. (2001). An object-oriented framework for distributed hydrologic and geomorphic modeling using triangulated irregular networks. *Computers & Geosciences*, *27*(8), 959–973. https://doi.org/10.1016/S0098-3004(00)00134-5 (cited on page 2).

van den Oord, A., Vinyals, O., & Kavukcuoglu, K. (2017). Neural discrete representation learning. *31st International Conference on Neural Information Processing Systems*, 6309–6318. https://doi.org/10.48550/arXiv.1711.00937 (cited on page 13).

Wang, X., Girshick, R., Gupta, A., & He, K. (2018). Non-local Neural Networks. *2018 IEEE/CVF Conference on Computer Vision and Pattern Recognition*, 7794–7803. https://doi.org/10.1109/CVPR.2018.00813 (cited on page 4).

Wu, J., Zhang, C., Xue, T., Freeman, W. T., & Tenenbaum, J. B. (2016). Learning a probabilistic latent space of object shapes via 3D generative-adversarial modeling. *30th International Conference on Neural Information Processing Systems*, 82–90. https://doi.org/10.48550/arXiv.1610.07584 (cited on page 3).

Xiao, Z., Kreis, K., & Vahdat, A. (2022). Tackling the Generative Learning Trilemma with Denoising Diffusion GANs. *10th International Conference on Learning Representations*. https://doi.org/10.48550/arXiv.2112.07804 (cited on page 13).

Xu, B., Wang, N., Chen, T., & Li, M. (2015). *Empirical Evaluation of Rectified Activations in Convolutional Network*. arXiv: 1505.00853 [cs, stat]. https://doi.org/10.48550/arXiv.1505.00853 (cited on page 3).

Xu, W. (1996). Conditional curvilinear stochastic simulation using pixel-based algorithms. *Mathematical Geology*, *28*(7), 937–949. https://doi.org/10.1007/BF02066010 (cited on page 1).

Zheng, Q., & Zhang, D. (2022). Digital Rock Reconstruction with User-Defined Properties Using Conditional Generative Adversarial Networks. *Transport in Porous Media*, *144*(1), 255–281. https://doi.org/10.1007/s11242-021-01728-6 (cited on pages 4, 12).

Zhou, Q., Yang, X., Zhang, R., Hosseini, S. A., Ajo-Franklin, J. B., Freifeld, B. M., Daley, T. M., & Hovorka, S. D. (2020). Dynamic Processes of CO Storage in the Field: 1. Multiscale and Multipath Channeling of CO Flow in the Hierarchical Fluvial Reservoir at Cranfield, Mississippi. *Water Resources Research*, *56*(2), e2019EF001360. https://doi.org/10.1029/2019WR025688 (cited on page 1).

# Appendix A  Further architecture and hyper-parameter tuning

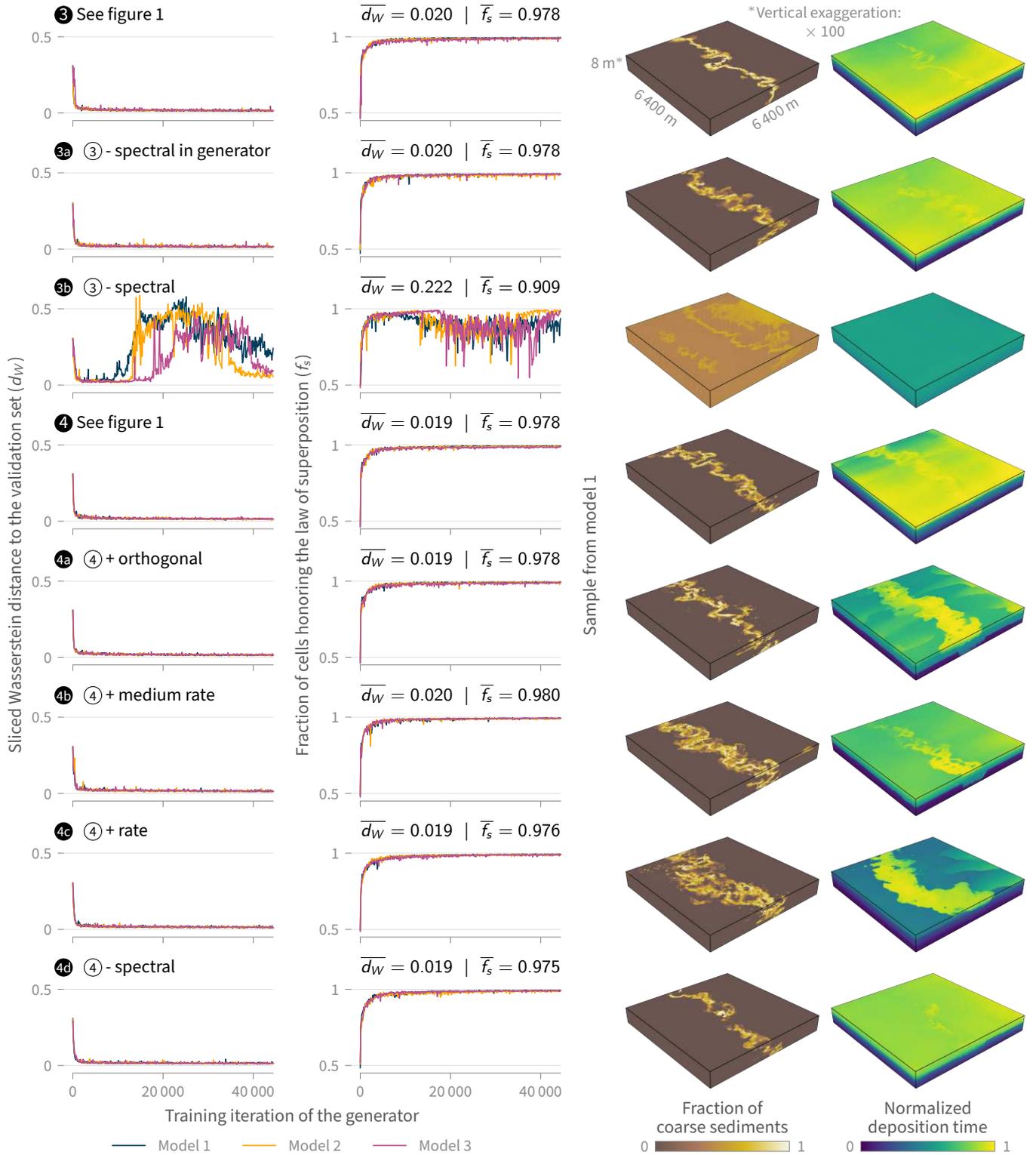

**Figure A.1**  Study starting from architectures 3 and 4 (figure 1) and testing how modifying some components affects training and sample quality. The samples come from the same latent vector. *spectral in generator*: spectral normalization in the generator only; *spectral*: spectral normalization in the generator and discriminator; *orthogonal*: orthogonal instead of normal initialization in the convolutional layers; *medium rate*: learning rate of 0.0001 for the generator and 0.0004 in the discriminator instead of 0.0002 for both; *rate*: learning rate of 0.00005 for the generator instead of 0.0002.

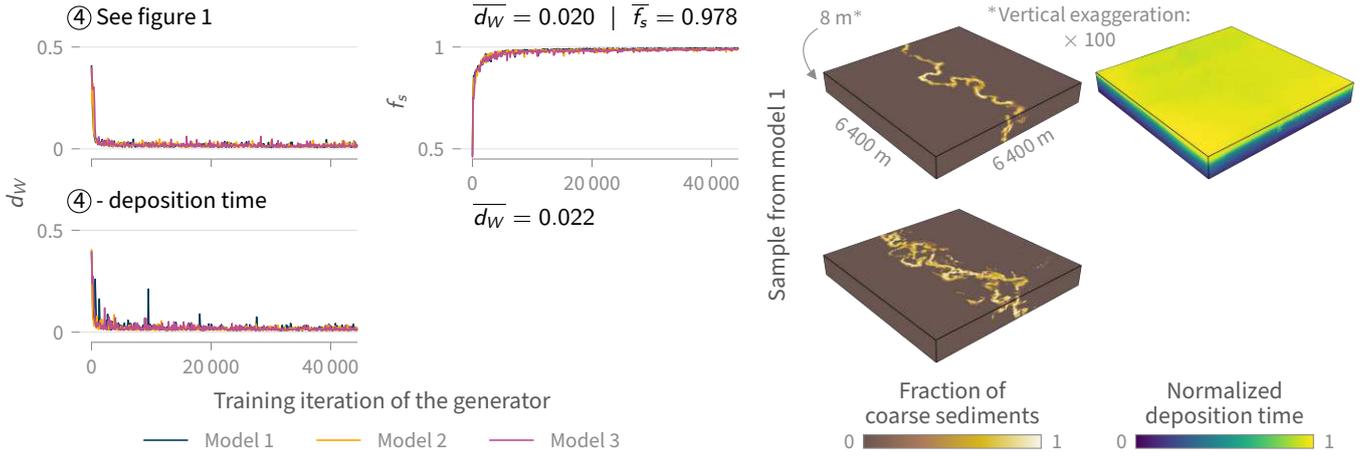

**Figure A.2** Impact of not generating the deposition time on training and sample quality of architecture 4 (figure 1). The samples come from the same latent vector. $d_W$: sliced Wasserstein distance to the validation set, which does not include the deposition time even in the top case; $f_s$: fraction of cells honoring the law of superposition; *deposition time*: samples only contain the fraction of coarse sediments and not the deposition time, so the law of superposition cannot be used for validation.

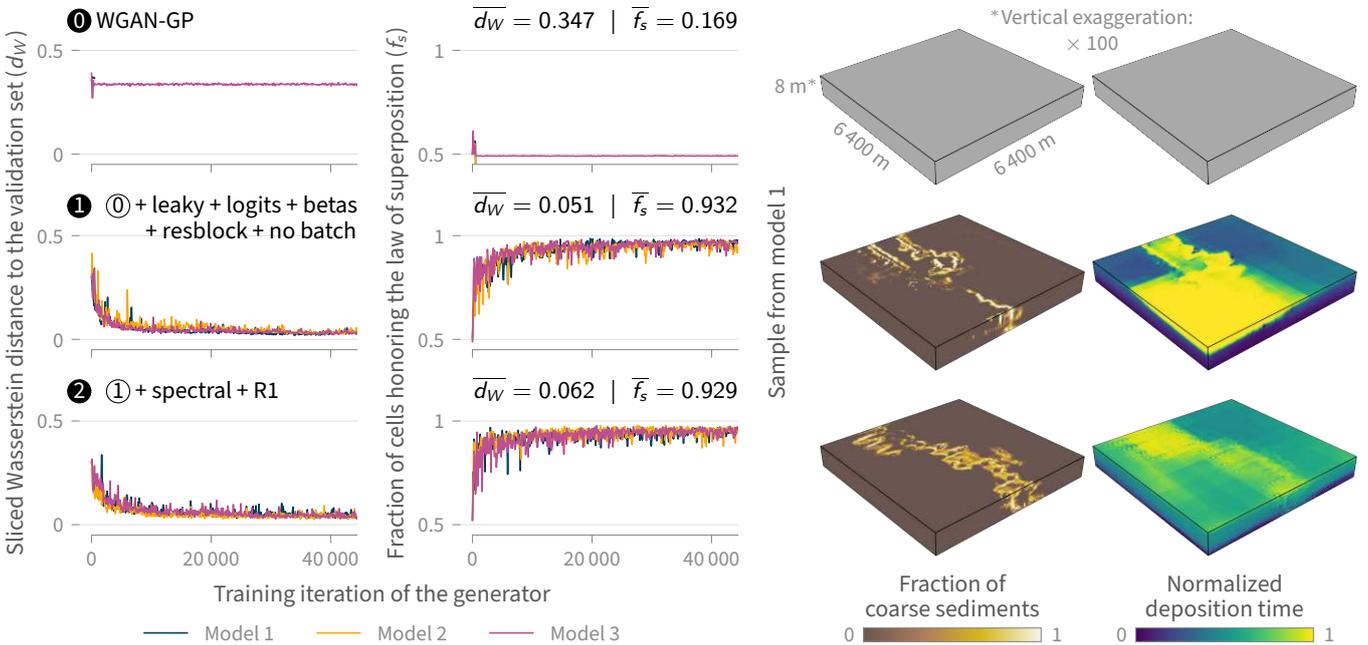

**Figure A.3** Ablation study starting from WGAN-GP and progressively adding elements of BigGAN to assess their effect on training and sample quality. The samples come from the same latent vector. *leaky*: Leaky ReLU in the generator instead of ReLU; *logits*: binary cross entropy with logits as loss instead of binary cross entropy with a sigmoid function as last activation in the discriminator; *betas*: $\beta_1$ of 0 and $\beta_2$ of 0.99 instead of 0.5 and 0.999; *resblock*: residual blocks in the generator and discriminator instead of the convolutional layers; *no batch*: no batch normalization in the discriminator; *spectral*: spectral normalization in the generator and discriminator; *R1*: $R_1$ regularization; *rate*: learning rate of 0.00005 for the generator instead of 0.0002.

# Appendix B  Further testing of a parsimonious architecture

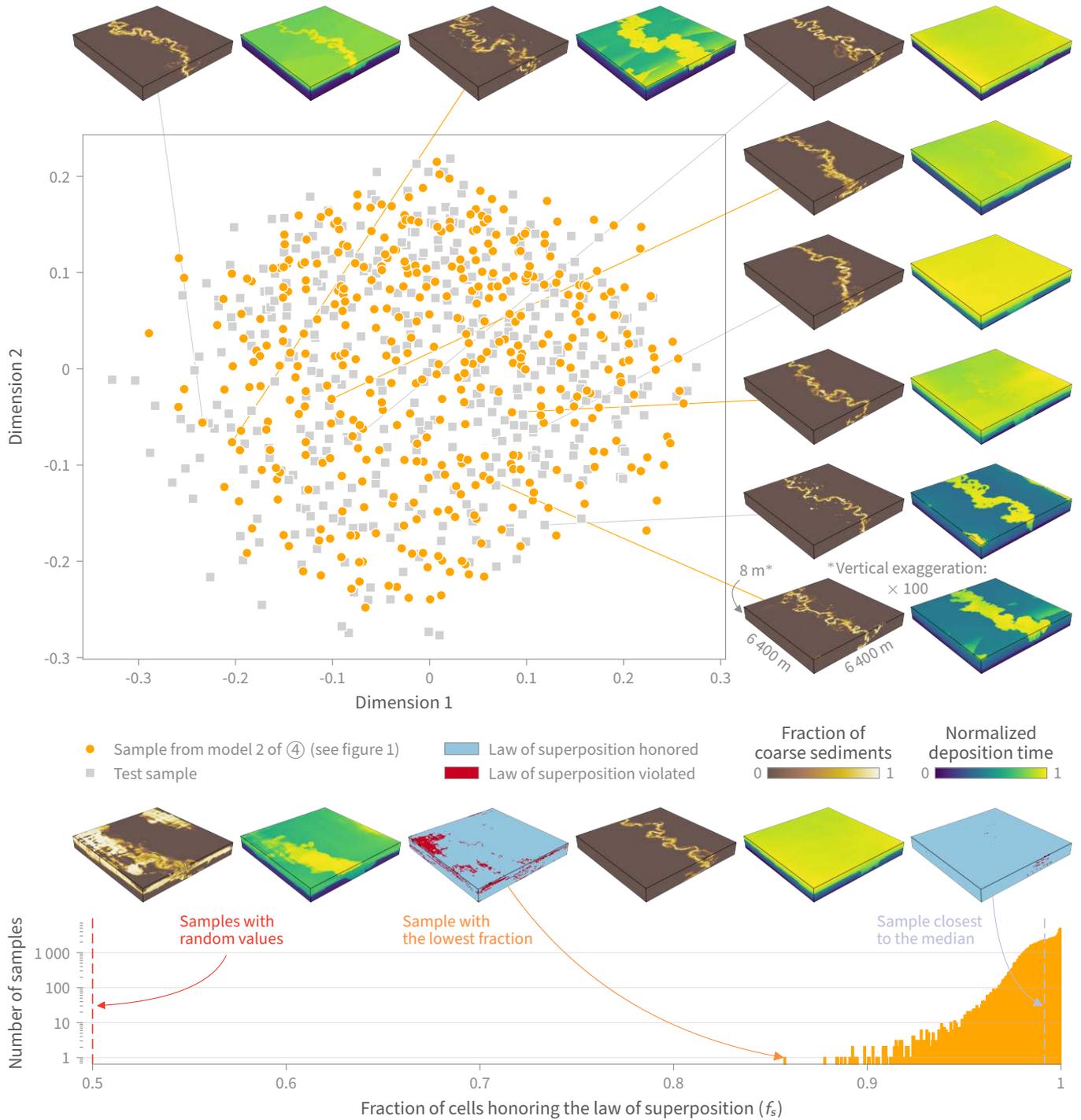

**Figure B.1**  Testing sample quality and diversity from model 2 of architecture 4 from figure 1 using multidimensional scaling to represent the sliced Wasserstein distances between 400 test samples and 400 samples from model 2 and the fraction of cells honoring the law of superposition in 10 000 samples from model 2.

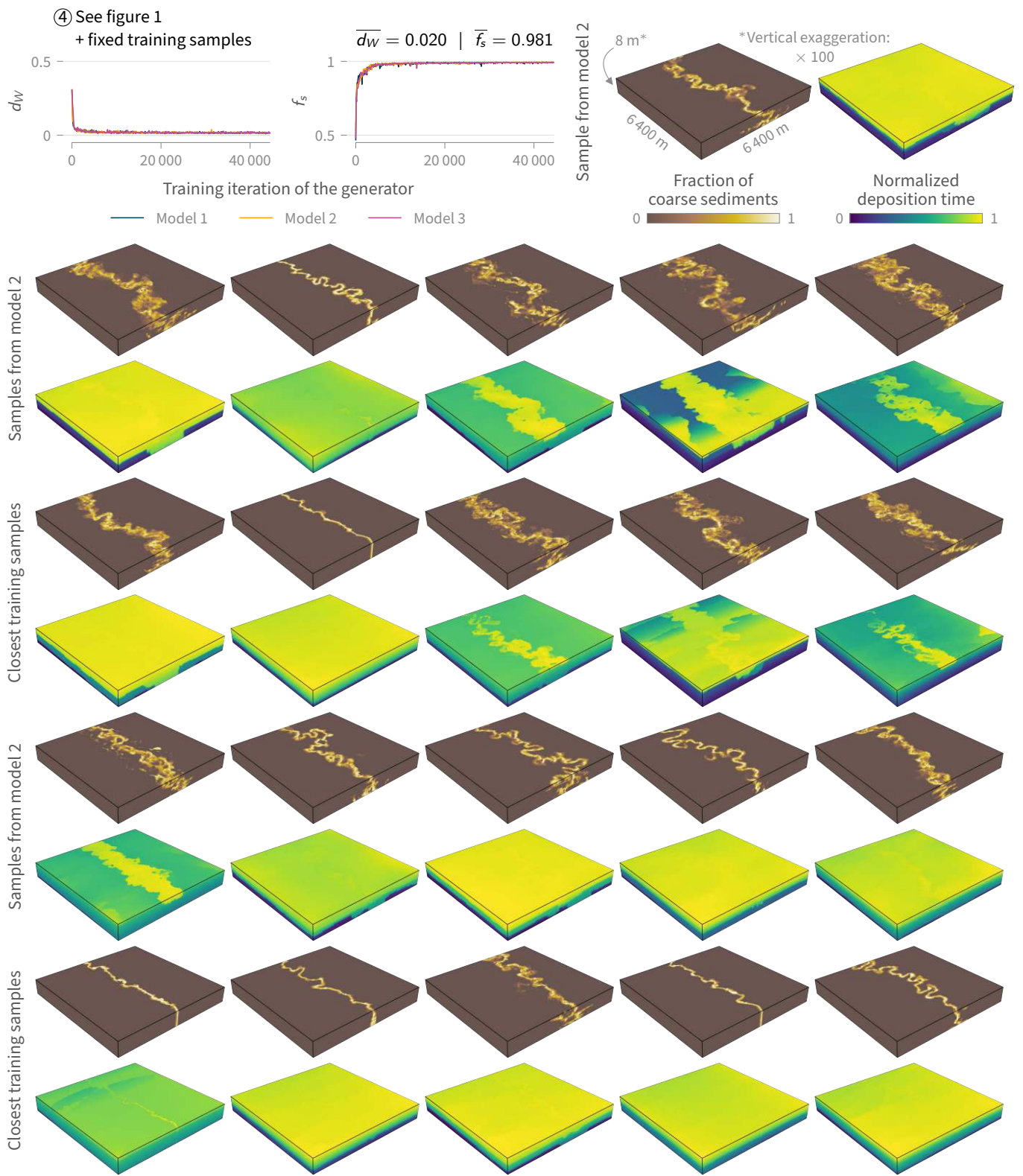

**Figure B.2** Testing memorization of model 2 of architecture 4 from figure 1 using a training with fixed instead of random samples extracted from FluvDepoSet. $d_W$: sliced Wasserstein distance to the validation set; $f_s$: fraction of cells honoring the law of superposition.

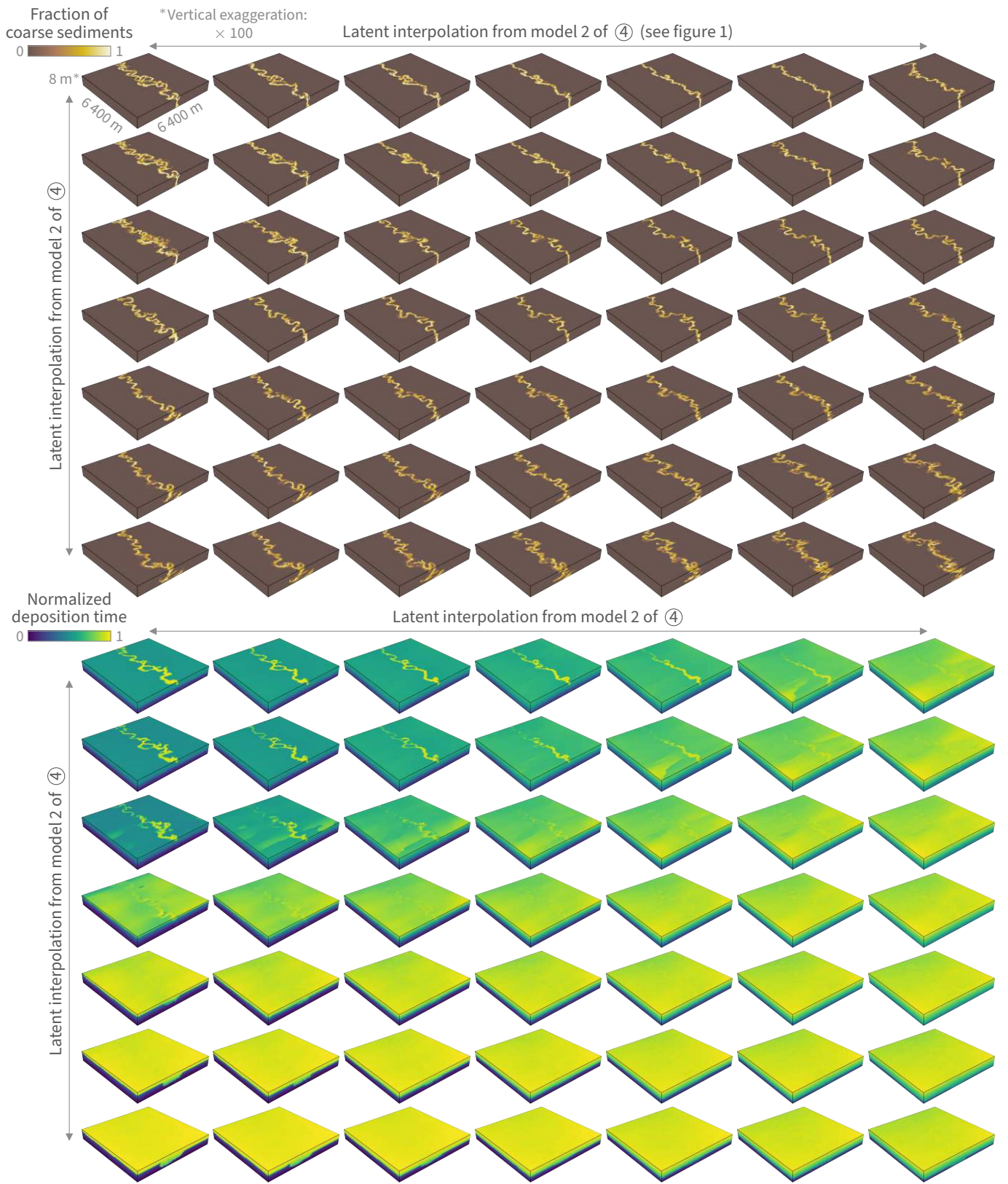

**Figure B.3** Latent interpolation between four samples (at each corner) of model 2 of architecture 4 from figure 1.

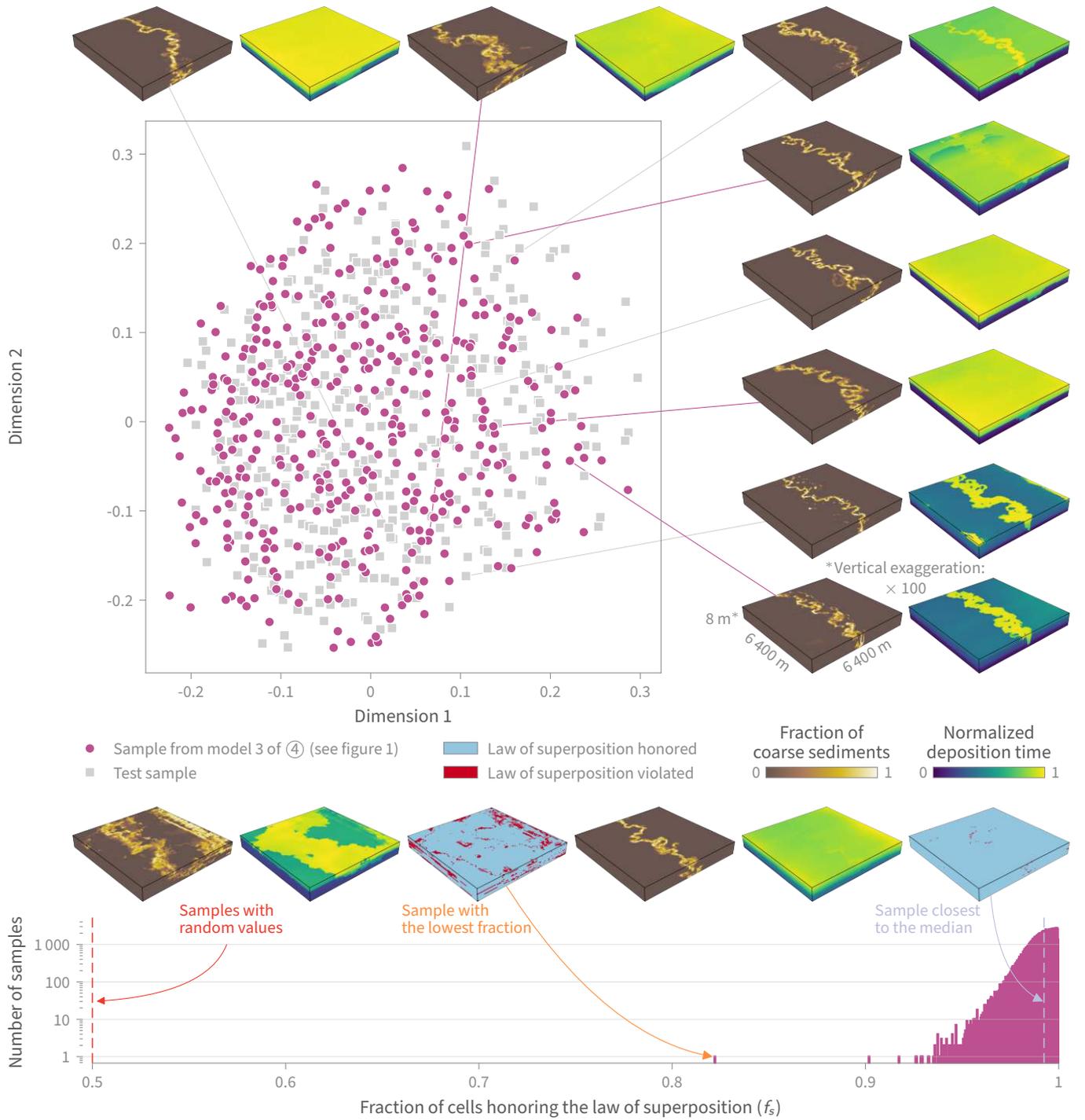

**Figure B.4** Testing sample quality and diversity from model 3 of architecture 4 from figure 1 using multidimensional scaling to represent the sliced Wasserstein distances between 400 test samples and 400 samples from model 3 and the fraction of cells honoring the law of superposition in 10 000 samples from model 3.

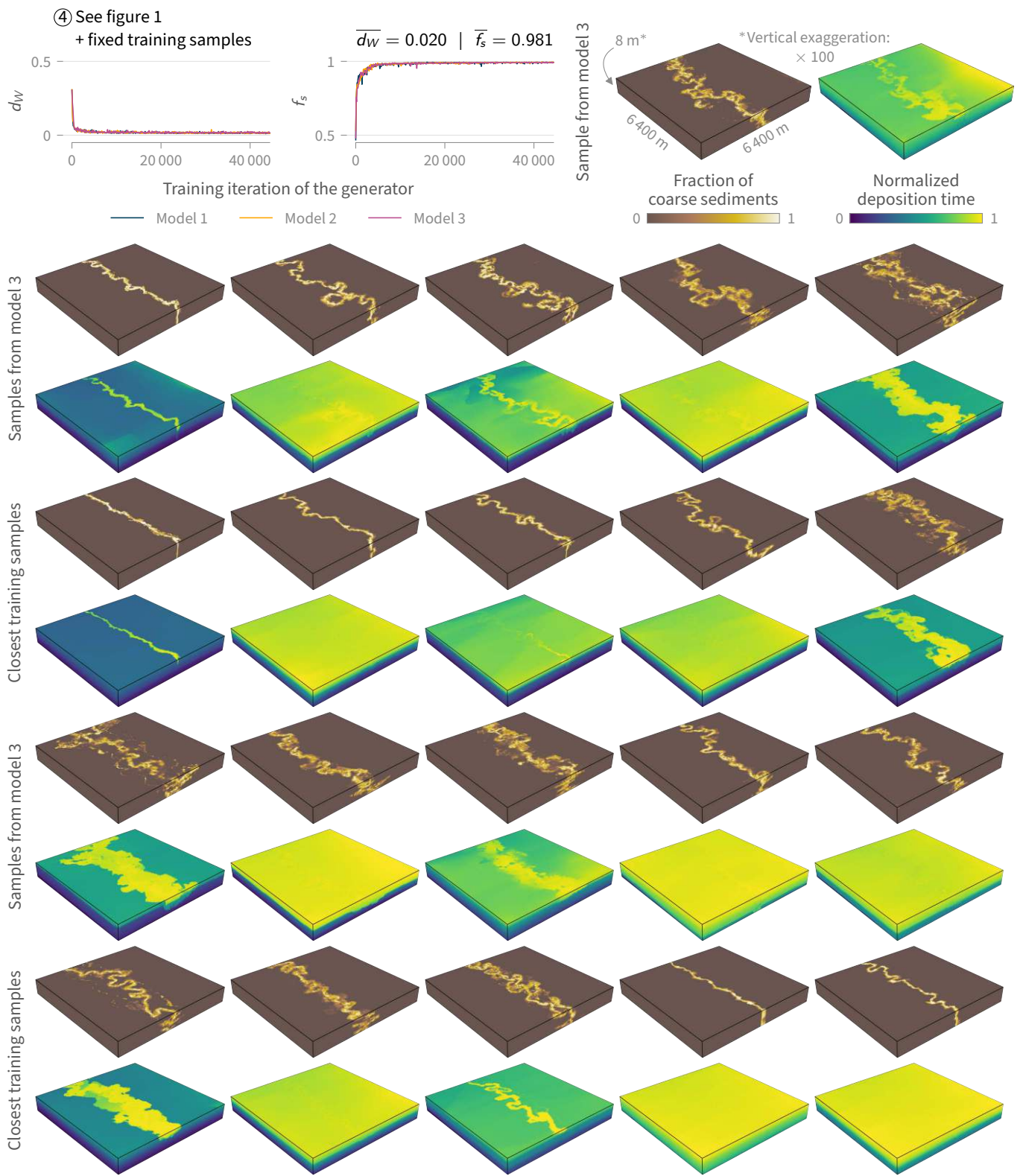

**Figure B.5** Testing memorization of model 3 of architecture 4 from figure 1 using a training with fixed instead of random samples extracted from FluvDepoSet. $d_W$: sliced Wasserstein distance to the validation set; $f_s$: fraction of cells honoring the law of superposition.

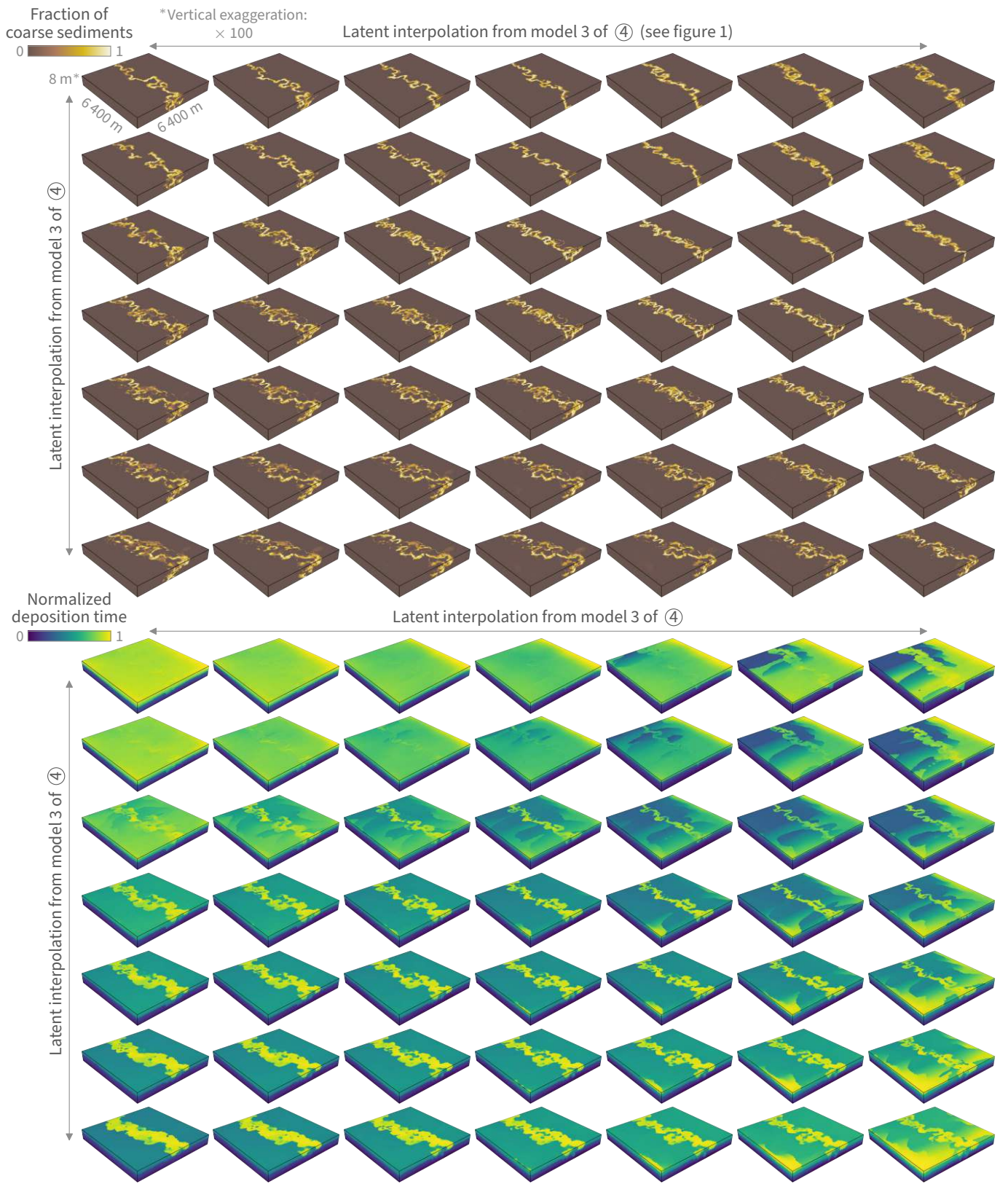

**Figure B.6** Latent interpolation between four samples (at each corner) of model 3 of architecture 4 from figure 1.